\documentclass[runningheads]{llncs}

\usepackage[mobile]{eccv}

\usepackage{eccvabbrv}

\usepackage{graphicx}
\usepackage{booktabs}

\usepackage{amsmath}
\usepackage{amssymb}
\usepackage{mathtools}
\usepackage{enumitem}
\usepackage{color}
\usepackage[stable]{footmisc}
\usepackage{tablefootnote}

\usepackage{gensymb}
\usepackage{stfloats}

\usepackage{csquotes}
\usepackage{bm}
\usepackage{float}
\usepackage{multirow}
\usepackage{tikz}
\usepackage[accsupp]{axessibility}  

\usepackage[breaklinks,colorlinks,citecolor=eccvblue]{hyperref}

\usepackage[capitalize]{cleveref}
\crefname{section}{Sec.}{Secs.}
\Crefname{section}{Section}{Sections}
\Crefname{table}{Table}{Tables}
\crefname{table}{Tab.}{Tabs.}

\usepackage{orcidlink}

\begin{document}

\title{RoScenes: A Large-scale Multi-view 3D Dataset for Roadside Perception}

\titlerunning{RoScenes}

\author{Xiaosu Zhu\inst{1}\thanks{Equal contribution. Work done when Xiaosu Zhu interns at Alibaba Cloud.} \and
Hualian Sheng\inst{1}\footnotemark[1] \and
Sijia Cai\inst{1}\thanks{Project lead.} \and
Bing Deng\inst{1} \and
Shaopeng Yang\inst{1} \and
Qiao Liang\inst{1} \and
Ken Chen\inst{2} \and
Lianli Gao\inst{3} \and
Jingkuan Song\inst{4}\thanks{Corresponding authors.} \and
Jieping Ye\inst{1}\footnotemark[3]}

\authorrunning{X.~Zhu et al.}

\institute{
Alibaba Cloud \and
Sichuan Digital Transportation Technology Co., Ltd \and
Independent Researcher \and
Tongji University \\
\email{\{zhuxiaosu.zxs, hualian.shl, stephen.csj\}@alibaba-inc.com}, \email{jingkuan.song@gmail.com}, \email{yejieping.ye@alibaba-inc.com}}

\maketitle

\begin{abstract}
  We introduce RoScenes, the largest multi-view roadside perception dataset, which aims to shed light on the development of vision-centric Bird's Eye View (BEV) approaches for more challenging traffic scenes. The highlights of RoScenes include significantly large perception area, full scene coverage and crowded traffic. More specifically, our dataset achieves surprising 21.13M 3D annotations within 64,000 $m^2$. To relieve the expensive costs of roadside 3D labeling, we present a novel BEV-to-3D joint annotation pipeline to efficiently collect such a large volume of data. After that, we organize a comprehensive study for current BEV methods on RoScenes in terms of effectiveness and efficiency. Tested methods suffer from the vast perception area and variation of sensor layout across scenes, resulting in performance levels falling below expectations. To this end, we propose RoBEV that incorporates feature-guided position embedding for effective 2D-3D feature assignment. With its help, our method outperforms state-of-the-art by a large margin without extra computational overhead on validation set. Our dataset and devkit will be made available at \url{https://github.com/xiaosu-zhu/RoScenes}.
\end{abstract}

\renewcommand{\arraystretch}{0.8}

\begin{figure}[t]
    \centering
    \includegraphics[width=\linewidth]{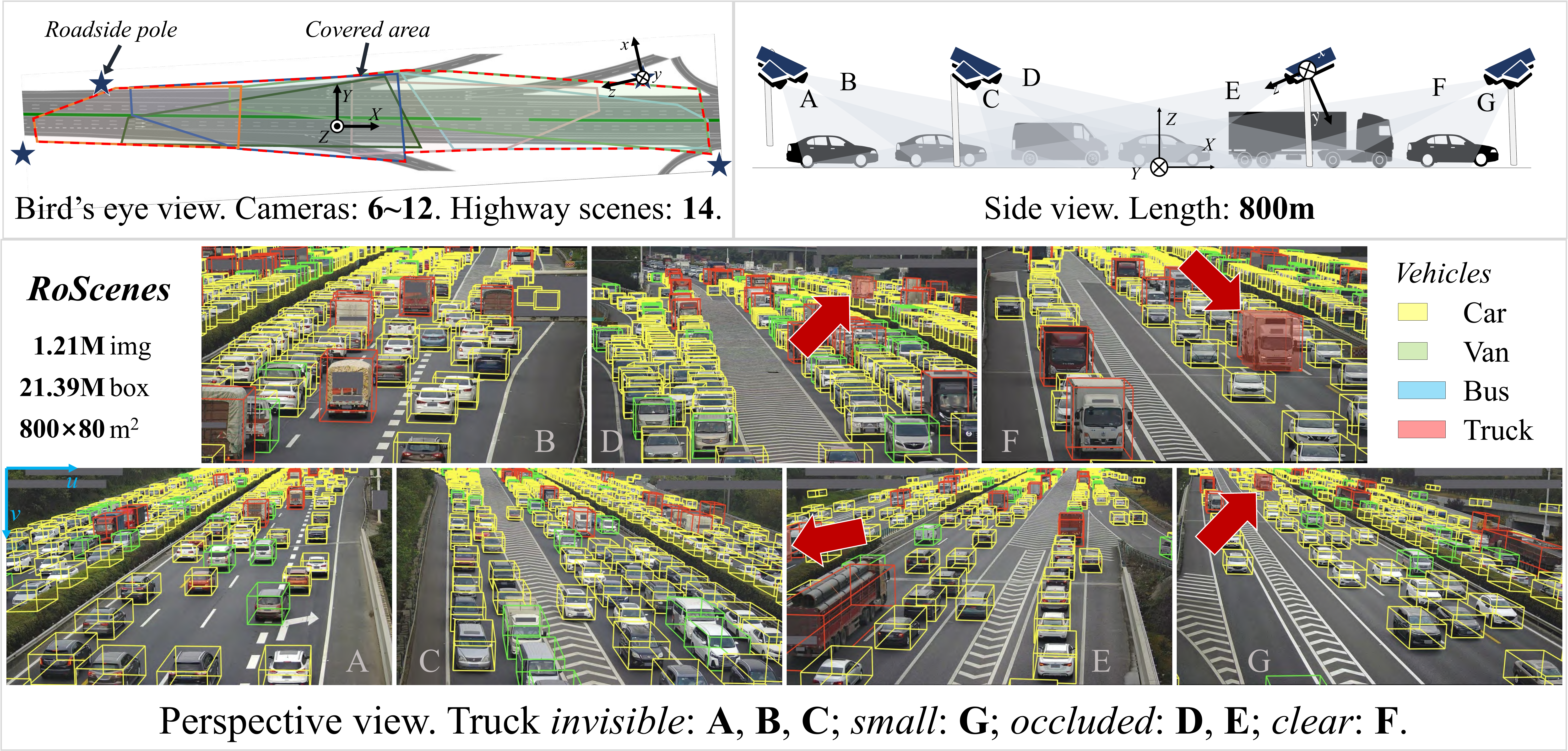}
    \caption{Demonstration of our RoScenes dataset. The annotated truck is difficult to recognize in A, B, C, E, F, G, but is clear in D.}
    \label{Fig.Hero}
\end{figure}

\begin{figure}[t]
    \centering
    \includegraphics[width=0.7\textwidth]{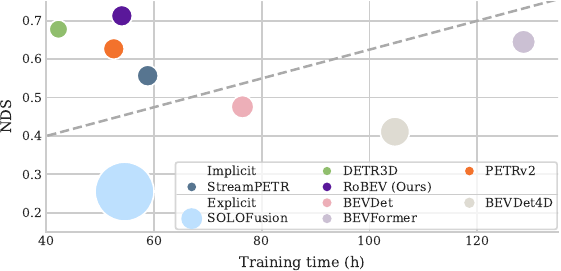}
    \caption{Performance (NDS), training time and inference model size comparison among two types of methods.}
    \label{Fig.Performance}
\end{figure}

\begin{table}[t]
\centering
\caption{Quantitative comparison with the published vehicle-side and infrastructure-side 3D datasets. Our dataset achieves the largest BEV perception area and the largest number of annotations. Type: \textbf{V}: Vehicle-side sensors. \textbf{I}: Infrastructure-side sensors. \enquote{Cam} is the number of synchronized cameras adopted per scene.}
\label{Tab.Main}
\resizebox{\linewidth}{!}{
\begin{tabular}{@{}lcccccrccrrr@{}}
\toprule
  \multicolumn{1}{c}{\multirow{2}{*}{Dataset}} &
  \multicolumn{1}{c}{\multirow{2}{*}{Year}} &
  \multicolumn{2}{c}{Type} &
  \multicolumn{1}{c}{\multirow{2}{*}{Cam}} &
  \multicolumn{1}{c}{\multirow{2}{*}{\begin{tabular}[c]{@{}c@{}}BEV area\\
  ($m^2$)\end{tabular}}} &
  \multicolumn{1}{c}{\multirow{2}{*}{\begin{tabular}[c]{@{}c@{}}Duration\\
  (\textit{hour})\end{tabular}}} &
  \multicolumn{2}{c}{Diversity} &
  \multicolumn{1}{c}{\multirow{2}{*}{Image}} &
  \multicolumn{1}{c}{\multirow{2}{*}{Box}} &
  \multicolumn{1}{c}{\multirow{2}{*}{Class}} \\ \cmidrule(lr){3-4} \cmidrule(lr){8-9}
     &      & \multicolumn{1}{c}{V} & \multicolumn{1}{c}{I} &   &   &    & Night & Rain &        &      &   \\ \midrule
KITTI~\cite{KITTI}&$2012$&\checkmark& - &$2$&$70\times80$&$1.5$&\checkmark&-&$0.02M$&$0.08M$&$8$\\
ApolloScape~\cite{ApolloScape}&$2019$&\checkmark& - &$6$&-&$2.5$&\checkmark&-&$0.14M$&$0.07M$&$8$-$35$ \\
nuScenes~\cite{nuScenes}&$2019$&\checkmark&-&$6$&$100\times100$&$5.5$&\checkmark&\checkmark&$1.40M$&$1.40M$&$23$\\
Argoverse~\cite{Argoverse}&$2020$&\checkmark& - & $7$ &$205\times155$&$320.0$&\checkmark&\checkmark&$0.02M$&$0.99M$&$15$ \\
Waymo Open~\cite{Waymo}&$2020$&\checkmark& - &$5$ &$150\times150$&$6.4$&\checkmark&\checkmark&$0.23M$&$12.00M$&$4$\\
ONCE~\cite{ONCE}&$2021$&\checkmark& - &$7$ &$200\times200$&$144.0$&\checkmark&\checkmark&$7.00M$&$0.42M$&$5$\\\midrule
Rope3D~\cite{Rope3D}&$2022$& - &\checkmark& $1$ &$104\times102$& - &\checkmark&\checkmark&$0.05M$&$1.50M$&$12$ \\
V2X-Seq~\cite{V2X-Seq}&$2023$&\checkmark&\checkmark& $1\text{+}1$ &$104\times102$&$0.4$&-&\checkmark&$0.07M$&$1.20M$&$10$ \\
A9~\cite{A9}&$2023$&-&\checkmark& $4$ &-&$0.1$&-&\checkmark&$0.01M$&$0.21M$&$9$ \\\midrule
\textbf{RoScenes} (Ours) & -    & - &\checkmark&$\mathbf{6\!\sim\!12}$&$\mathbf{800\times80}$&$23.9$&\checkmark&-&$\mathbf{1.30M}$&$\mathbf{21.13M}$&$4$ \\
\bottomrule
\end{tabular}
}
\end{table}

\section{Introduction}
\label{Sec.Intro}
3D roadside perception is one of the recent trends in the field of intelligent transportation systems (ITS). It has garnered significant attention for its potential applications in innovative traffic solutions, such as the digital twin of road traffic and cooperative vehicle infrastructure systems~\cite{DAIR-V2X,V2X-Seq}. Nevertheless, the research progress in 3D roadside perception has lagged behind other domains such as autonomous driving (AD), primarily due to the lack of large-scale nuScenes-like benchmarks~\cite{nuScenes} and sophisticated roadside configurations for wide-range 3D perception algorithms. To remedy these defects and conduct pilot studies on a variety of 3D roadside perception tasks, in this paper, we introduce the largest multi-view perception dataset for \textbf{\textit{Ro}}adside \textbf{\textit{Scenes}}, called \textbf{\textit{RoScenes}}. Our dataset aims at extending the recent advanced perception frameworks (\eg, vision-centric BEV perception methods~\cite{BEVFormer,LSS,PETR,StreamPETR,BEVDet,DETR3D,Sparse4D}) for the real-world roadside challenges such as cross-camera fusion, poor 3D localization and heavy traffic monitoring.

As shown in \cref{Fig.Hero}, RoScenes focuses on roadside 3D object detection task and has several appealing properties: 1) \textbf{Large Perception Range}: It contains $14$ highway scenes, each captured by $6 \sim 12$ cameras in a rectangle field of $800\times80 m^2$ that is $\sim\!6\times$ larger than vehicle/roadside datasets (\eg, nuScenes~\cite{nuScenes} or Rope3D~\cite{Rope3D}). To the best of our knowledge, RoScenes has the largest sensing range among real-world traffic datasets with accurate 3D annotations. Such a broad perception range has the potential to enable precise measurement of vehicle trajectories to forecast safety-critical highway situations. 2) \textbf{Full Scene Coverage}: The conventional sensor placement scheme in current roadside datasets is isolated and sparse~\cite{Rope3D,DAIR-V2X,V2X-Seq,A9}, resulting in blind spots and has intrinsic limitations to be utilized for mutli-view, high-performance 3D object detection. Our roadside cameras of each scene are mounted on $4 \sim 6$ different poles with high position, various pitch angles and focal lengths. These camera configurations mitigate dynamic occlusions and have significant overlapping coverage to leverage multi-view geometry. For instance, the annotated vehicle in \cref{Fig.Hero} is hard to distinguish in some views but is still quite clear in at least one view. 3) \textbf{Crowded Traffic Scenes}: To promote new studies on the unsolved challenges of crowded traffic environments, especially for query-based 3D object detection methods, RoScenes also collects a large amount of camera frames with highly dense and obstructed vehicles.
The maximum number of vehicles per scene sample and per camera view are $567$ and $293$. The dense crowds with severe occlusions make most approaches difficult to infer the existences or accurate 3D positions of vehicles. Therefore, a computational effective architectures are in great need and of practical importance. We highlight the characteristic of RoScenes as the essentially different BEV setup and the multiple world-to-cameras geometry relations other than a specific camera setup and layout, compared to existing autonomous driving datasets.

The 3D annotation pipeline of RoScenes is another contribution. Recent realistic roadside datasets in labeling 3D objects mostly rely on LiDAR sensors, which are very expensive and have poor stability in
challenging
roadside conditions. In addition, the annotation process of LiDAR-Camera is cumbersome and still suffers from quality concerns in the context of congested scenes. To address these practical problems, we propose a BEV-to-3D joint annotation pipeline based on a pre-built 3D scene reconstruction model and time-synchronized image data among roadside cameras and Unmanned Aerial Vehicles (UAVs). We first utilize the offboard calibration techniques to obtain reliable intrinsic and extrinsic parameters of cameras, as well as the UAV-to-World homography parameters. Our pipeline starts with a pretrained BEV 2D detector and tracker to predict the BEV 2D detections and tracking IDs. Then the 2D boxes are converted to world 3D boxes based on homography transformation and class-fixed height lifting. In order to achieve accurate sensor synchronization, we use camera parameters to reproject the 3D boxes into each scene and perform projective and temporal alignment. Most steps of our pipeline are implemented automatically to accelerate the annotation process. For a new roadside scene, the overall preparation is within $3h$ ($2h$ for 3D reconstruction, $1h$ for calibration, synchronization and refinement). Then, we could produce up to 20k samples per day, while a skilled worker can only annotate $100 \sim 200$\cite{ONCE}. We finally generates $21.13$ million 3D boxes for $1.30$ million roadside images in RoScenes.

Due to the large quantity of vehicle annotations and collaborative camera layout, RoScenes poses significant challenges to 3D roadside perception. One major challenge is the cross-view object association problem in traditional multi-view late fusion paradigm, suffering from poor generalization performance in complex scenes. To explore the capability of the model in leveraging multi-view complementary features, we adopt the popular BEV detection methods without additional stitching strategies in overlapping areas. To this end, we explore explicit~\cite{BEVFormer,BEVDet,BEVDet4D,SOLOFusion} as well as implicit~\cite{DETR3D,PETRv2,StreamPETR} BEV methods and report results in \cref{Fig.Performance}, where issues exist in both type of methods. Explicit ones consume a high computational cost, while implicit ones suffer from ineffective 2D-3D interaction raised by variation of camera layouts. To tackle these challenges, we propose RoBEV for RoScenes which is built upon implicit paradigm for efficiency, and incorporates feature-guided 3D position embedding for effective 2D-3D feature assignment. We conduct extensive experiments and make comprehensive study for various methods on RoScenes, and results show that our RoBEV outperforms state-of-the-art methods by a large margin.

In summary, the contributions of our work include:
1) We release a large-scale multi-view 3D dataset for roadside perception. Meanwhile, a novel, cost-effective annotation pipeline is introduced to obtain 3D annotations in challenging traffic scenarios. It will be made publicly available to the research community, and we hope it will promote the development of advanced roadside models.
2) We design RoBEV that effectively aggregates 2D image feature to 3D detection queries via feature-guided position embedding. Our method achieves superior performance over the state-of-the-art BEV detection approaches on the RoScenes dataset. 3) The extensive experimental evaluation also indicates the RoScenes dataset can serve as a benchmark for BEV architectures in the future.

\section{Related Works}
\label{Sec.RelatedWorks}
Detecting traffic participants has drawn high attention in AD and ITS areas. In this section, we briefly review the related datasets and BEV approaches.

\noindent\textbf{Vehicle-side/Infrastructure-side Datasets.}
There exists numerous 3D datasets that offer image sequences captured by driving scenarios, along with dense 3D annotations. These datasets include
KITTI~\cite{KITTI}, ApolloScape~\cite{ApolloScape}, H3D~\cite{H3D}, nuScenes~\cite{nuScenes}, A*3D~\cite{A3D}, A2D2~\cite{A2D2}, Argoverse~\cite{Argoverse}, Waymo Open~\cite{Waymo}, \etc.
Naturally, since sensors are mounted on cars, the field of view is
focused on the horizon,
resulting in frequent occlusion of distant objects.

Several recent works have proposed a solution by utilizing roadside cameras~\cite{A9,Rope3D,DAIR-V2X,V2X-Seq}. Roadside cameras are lifted over the ground to alleviate occlusion and obtain far field of view. Therefore, these sensors are more suitable for long-range long-term perception. Works like DAIR-V2X~\cite{DAIR-V2X} and V2X-Seq~\cite{V2X-Seq} record roadside and vehicle-side images cooperatively, while Rope3D~\cite{Rope3D} and A9~\cite{A9} provide pure roadside data. However, current works still locate in a relatively small area with a few cameras and LiDARs, which limits the perception capacity of roadside systems. Therefore, a multi-view roadside perception dataset is needed to enlighten the research and industrial application.

\noindent\textbf{BEV-based Multi-view 3D Object Detection.} The vision-centric BEV perception framework.can be divided into two categories: explicit and implicit approaches. Explicit ones involve a BEV feature map for prediction. For instance,
Lift-splat~\cite{LSS,BEVDet} is a pioneering method that
provides
a practical way to aggregate 2D image features into 3D BEV
using the estimated depth distribution.
Following works~\cite{BEVDepth,BEVDet4D} explore depth supervision or temporal fusion to refine BEV features. Transformer-based methods~\cite{BEVFormer,BEVFormerV2} further employ the attention mechanism~\cite{Attention} to perform aggregation.

In contrast, implicit methods directly make predictions from inputs. These methods follow the idea of DETR~\cite{DETR} to learn a set of detection queries to extract features from 2D images by attention.
Typical methods include DETR3D~\cite{DETR3D}, SparseBEV~\cite{SparseBEV} and PETR series~\cite{PETR,PETRv2,StreamPETR}. The first two sample from local image features into queries based on 2D-3D projection, while the latter attach position-aware embedding on image features for global feature assignment. 
These methods incur lower costs and offer faster inference speed as they do not require construction and computation over the entire BEV feature map.

\begin{figure}[t]
    \centering
    \includegraphics[width=\linewidth]{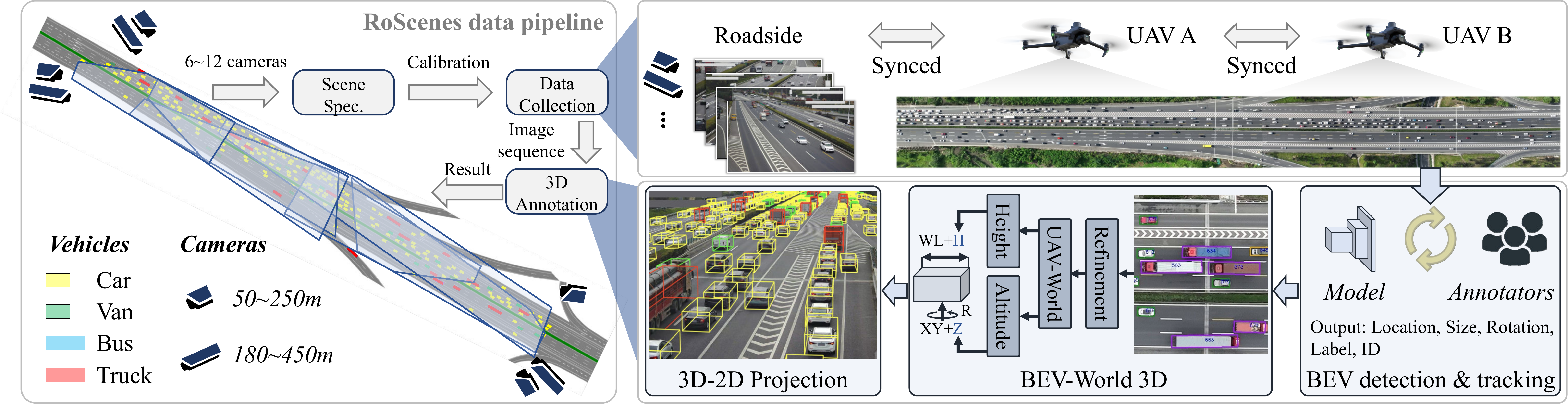}
    \caption{Overall data collection and annotation pipeline. We propose BEV-to-3D joint annotation for efficiency.}
    \label{Fig.Pipeline}
\end{figure}

\section{The RoScenes Dataset\footnote{Scene samples, trajectories visualization and more analysis appear in appendix.}}

In this section, we describe the scene specification, data pipeline, statistics and evaluation protocol of RoScenes.



\subsection{Scene Specification}
\label{Sec.Spec}
\textbf{Task Setup.} RoScenes is primarily created for multi-view 3D object detection task and we also provide the monocular 3D object detection setup in appendix for the broader research community.

\noindent\textbf{Scene Setup.} The whole dataset consists of $14$ highway scenes in the Chengdu Ring Expressway, Sichuan, China, which is known for its heavy traffic conditions. As shown in \cref{Fig.Hero}, a typical scene setup contains $4\sim 6$ poles placed alongside the inner ring road and outer ring road, with $6\sim 12$ cameras installed in total. The mounting location, height, and orientation of cameras are carefully adjusted to cover $\!\sim\!800$ meters without blind zones. We then build the real-texture 3D reconstruction model of each scene using UAV oblique photography and use it for the calibration. For data collection, we denote a scene sample as a group of images captured synchronously from all cameras in a scene. A clip consists of $60$ continuous scene samples at $2$Hz, which is the basic unit in the dataset.

\noindent\textbf{Sensor Setup.} Roadside cameras are mounted on the highway poles from a height of above $10$m. On each side of the pole, there are usually $2$ cameras with different zoom levels covering near-range and far-range vehicles, respectively. For data collection, we use $2$ UAVs flying at height of $300$m to scan each highway scene. We list the detailed sensor specifications as below:
\begin{itemize}
    \item Camera: Hikvision cameras, zoom lens, with 4k resolution, $25$Hz capture frequency and $1/1.2''$ CMOS sensor.
    \item UAV: DJI Mavic 3E with a wide camera, a tele camera and an RTK module for centimeter-level positioning. $20$MP wide-angle camera is with $30$Hz capture frequency and $4/3''$ CMOS sensor. $12$MP tele camera is with $30$Hz capture frequency and $1/2''$ CMOS sensor.
\end{itemize}

\noindent\textbf{Coordinate Systems and Calibration.}
To obtain reliable transformation matrices between different sensors, we first adopt the local Universal Transverse Mercator (UTM) coordinate system of the 3D scene reconstruction as our World Coordinate. Subsequently, adhering to the conventions of KITTI~\cite{KITTI}, we define the Camera and Image coordinates, as shown in \cref{Fig.Hero}.

Inspired by \cite{pytorch3d}, we exploit a joint estimation algorithm for camera intrinsic and Camera-to-World extrinsic parameters to achieve accurate camera calibration in the outdoor scenarios. Specifically, we first compute the initial parameters from the correspondences between the camera image and the 3D scene reconstruction. Then we apply the differentiable rendering to build an iterative scheme in which the intrinsic and extrinsic parameters are alternatively optimized. The final camera parameters are determined by the minimum reprojection error within a fixed iterations. In addition, the UAV-to-World calibration can be obtained by simply calculating the planar homography by flat-world assumption. We verify the 3D static reconstruction error via DJI Terra\footnote{\url{https://enterprise.dji.com/dji-terra}}, which reports $3.74cm$/$6.11cm$ absolute horizontal/vertical accuracy, respectively. The actual error is further verified to be $<10cm$ by GPS real-time kinematic positioning\cite{GPS-RTK} and Qianxun high-definition map\footnote{\url{https://www.qxwz.com/}} on several ground control points. Finally, we construct a standard BEV perception cuboid with $X\!\in\![-400, 400], Y\!\in\![-40, 40], Z\!\in\![0, 6]$ and translate scenes to the cuboid to remove sensitive UTM information for data privacy.

\noindent\textbf{Sensor Synchronization.} For the temporal calibration, we first synchronized the roadside cameras with a Network Time Protocol (NTP) time server. To reach accurate synchronization between cameras and UAVs, we then project the vehicle 3D boxes (obtained in \cref{Sec.Collection}) of continuous UAV frames into roadside cameras and select the best projection quality to determine the time shift.



\subsection{BEV-to-3D Joint Annotation}
\label{Sec.Collection}
The large amount of collected data makes any manual annotations impracticable. Thus, we design an efficient BEV-to-3D joint annotation pipeline which is mostly automatic in producing 3D boxes, IDs and class labels simultaneously. The general idea of our pipeline is to employ UAV for BEV annotation with no occlusion issues.

We present a schematic overview of our pipeline in \cref{Fig.Pipeline} including 4 key steps: 1) A couple of UAVs hover the target scene to capture aerial image sequence along with roadside camera sequence synchronously; 2) We train the UAV model consists of BEV detector and tracker on UAV images for generating image-level BEV annotations, which are then transformed to the XY plane of World Coordinate via the UAV-to-World homography matrix; 3) To further access the altitude and height of each annotated vehicle for converting BEV 2D boxes to world 3D boxes, we choose the center of BEV 2D boxes to query the altitude in pre-built 3D reconstruction model and attach vehicle's height to the average height of its class label; 4) We perform the perspective projection of 8 corners of 3D boxes onto 2D image planes for all roadside cameras using the camera parameters.

The annotation quality heavily depends on the BEV 2D detection models and 3D-to-2D projection. Therefore, we design a highly reliable model and a geometric refinement module for correcting BEV boxes.

\noindent\textbf{Design of BEV 2D Detector and Tracker.} We adopt the state-of-the-art aerial detection model RTMDet~\cite{RTMDet} pretrained on DOTA~\cite{DOTA} for BEV 2D box detection with location, size, rotation and class label. Then, a multi-object tracking algorithm OC-SORT~\cite{OCSORT} is applied upon detection results to produce trajectories for every passing vehicle. We achieve the very high detection \& tracking quality via progressive fine-tuning and post-processing in our scene. The former is done by repeatedly collecting bad cases and re-training the model. At that time, $24,484$ training images make detection mAP $\geq0.95$. Next, we refine tracking trajectories by short trajectory pruning and future frame interpolation. In summary, we report $623$ ($0.0086\%$) / $454$ ($0.0063\%$) false positives / false negatives and $24$ ($0.016\%$) ID switches, by manually checking on $7.26M$ boxes and $150k$ trajectories, respectively.\footnote{The check covers $30\%$ annotations from RoScenes and about $7k$ samples from unpublished $31$ scenes.} Meanwhile, we pick $40k$ boxes to compare 2D length and width error with human annotations. The error is less than $1.16$ pixel in UAV view, corresponding to $12.53cm$ in physical size.

\noindent\textbf{Refinement of BEV Annotations.} The predicted BEV boxes suffer from the perspective distortions and jittering effects of UAV, which degrade the accuracy of vehicle's length and 3D-to-2D projection. To alleviate these issues, we employ point feature matching to stabilize UAV images and the triangulation strategy to reduce the length error.

\noindent\textbf{Annotation Format.} Our 3D annotations provide location, size, orientation and clip-level tracking ID for all vehicles. We further label vehicles into $4$ classes: \texttt{car}, \texttt{van}, \texttt{bus} and \texttt{truck}.

\noindent\textbf{Data protection.} Before the public release, we would erase all visible plates on vehicles, mask sensitive information and traffic signs to ensure data privacy. Meanwhile, as described in the formulation of coordinate systems, there should be no access to real geographic information.


\subsection{Statistics and Analysis}
\label{Sec.Stat}
With the efficient data collection and annotation pipeline, we are able to equip the community with a large-scale and high-quality 3D object dataset from the real traffic scenarios, which can facilitate a variety of 3D roadside vision tasks and potential applications. As shown in \cref{Tab.Main}, our dataset includes the largest perception area and correspondingly the maximum amount of 3D annotation when compared to the existing AD and roadside datasets. Next, we describe the annotation distributions and analyze the properties of RoScenes.

\begin{figure}[t]
    \centering
     \begin{subfigure}[b]{0.18\linewidth}
         \centering
         \includegraphics[width=\linewidth]{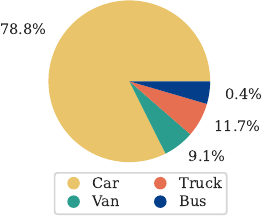}
     \end{subfigure}
     \hfill
     \begin{subfigure}[b]{0.8\linewidth}
         \centering
         \includegraphics[width=\linewidth]{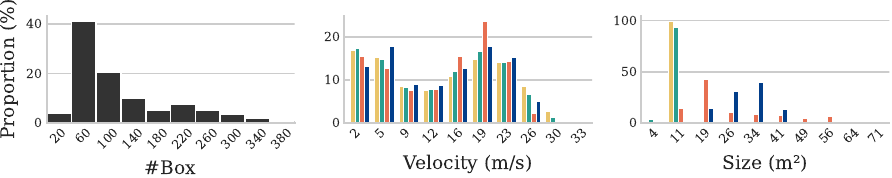}
     \end{subfigure}
    \caption{Summary of all 3D annotations. 1st: Pie chart of different vehicle types. 2nd: Histogram of box amount per scene sample. 3rd: Velocity statistics of different vehicles. 4th: Size (width$\times$height) statistics of different vehicles.}
    \label{Fig.Stat}
\end{figure}

\noindent\textbf{Annotation Statistics.} We first show the distributions of annotated vehicles in \cref{Fig.Stat}, including proportion of different vehicles, number of annotations per scene sample, vehicle velocity and size. Specifically, we observe that almost all passing vehicles are cars in this highway. Among annotated results, an average of $123$ boxes appear for every scene sample. While for every perspective image, this number is $71$, which is commonly $3\times$ larger than previous datasets (nuScenes: $9.7$, Rope3D: $24$). Two peaks appear at $60$ and $220$ in the 2nd figure and correspondingly $19m/s$, $5m/s$ in the 3rd. These are with the normal and congested traffic conditions respectively. The last column shows that trucks and buses occupy a larger variance of size (weight$\times$height) than cars and vans.

\begin{figure}[t]
    \centering
    \begin{subfigure}[t]{0.49\linewidth}
    \centering
        \includegraphics[width=\linewidth]{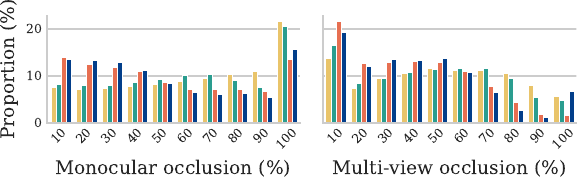}
        \caption{Monocular and multi-view occlusion.}
        \label{Fig.OCC}
    \end{subfigure}
    \begin{subfigure}[t]{0.49\linewidth}
    \centering
        \includegraphics[width=\linewidth]{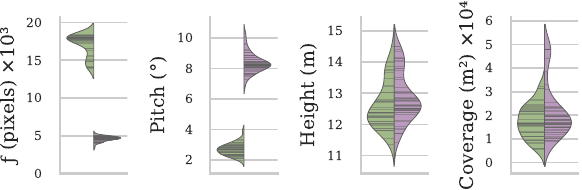}
        \caption{Violin plot of camera parameters.}
    \label{Fig.CameraParams}
    \end{subfigure}
    \caption{Camera statistics in terms of occlusion, focal length, pitch, mounting height and road coverage. Monocular/multi-view occlusions are grouped by vehicle types. Camera parameters are grouped by camera types. (green: far-range, purple: near-range)}
\end{figure}

\begin{figure}[t]
    \centering
    \includegraphics[width=0.7\linewidth]{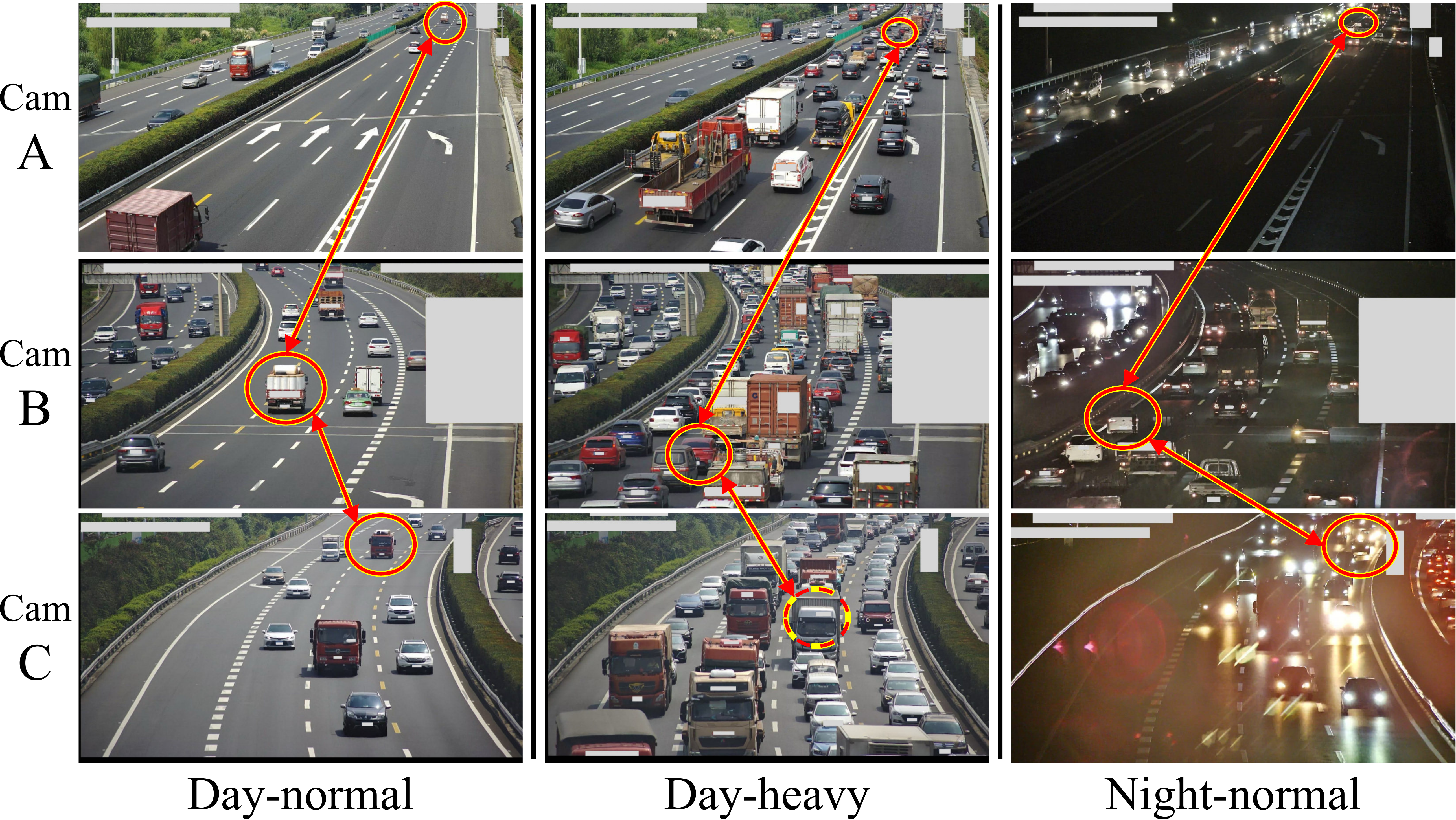}
    \caption{Multi-view images under different conditions. Connected vehicles are identical.}
    \label{Fig.Condition}
\end{figure}

\noindent\textbf{Scene Conditions.} As illustrated in \cref{Fig.Condition}, RoScenes contains 3 typical scene conditions: normal daytime traffic, heavy daytime traffic and normal night traffic (denoted as Day-normal,  Day-heavy and Night-normal). In Day-normal scenes (1st column), most vehicles can be easily distinguished with the benefit of multi-view camera setting. However, for Day-heavy (2nd column) scenes, a lot of cars are occluded by big trucks, making them hard to locate. The 3rd column of Night-normal scenes shows that almost all vehicles are hard to detect with extremely low light, emphasizing the need of illumination robustness in models.

\noindent\textbf{Sensor Layout Analysis.}
The complexity of the roadside environment results in varying camera layouts and setups across different scenes.
\cref{Fig.CameraParams} plots statistics of all cameras we use, where we can see the large variations in focal length, pitch, mounting height and coverage of road. Note that previous vehicle-side datasets mainly adopt similar or identical layout across scenes, RoScenes raises new challenges for current algorithms to encode the 2D-3D geometry priors of diverse scenes and generalize to novel scenes.

\noindent\textbf{Occlusion Analysis.} We take a study on monocular and multi-view occlusion and show the difference especially under our multi-view setting. For a scene sample, given an arbitrary perspective image of camera $c$, the monocular occlusion of a vehicle 3D box $\bm{b}_i$ is calculated as follows:
\begin{equation}
    \mathit{occ}_i^c=
    \begin{cases}
    \frac{\lVert\left\{\cup\hat{\bm{b}}^c_j\right\} \cap \hat{\bm{b}}^c_i\rVert}{ \lVert\hat{\bm{b}}^c_i\rVert},\; &\lVert\hat{\bm{b}}^c_i\rVert > 0, \bm{b}_j \in \bm{\Omega}^c_i\\
    1,\; &\mathit{otherwise},\\
    \end{cases}
\end{equation}
where $\hat{\cdot}^c$ is projected polygon of 3D box under perspective view $c$. We pick boxes that are nearer than $\bm{b}_i$ in $c$'s view (by calculating distance from box to $c$) as a set $\bm{\Omega}_i^c$. We union projected polygons of all boxes in $\bm{\Omega}_i^c$, and intersects it with $\hat{\bm{b}}_i^c$ to get $\bm{b}_i$'s occluded part in $c$'s view. $\lVert\cdot\rVert$ is the area of polygon. The calculated $\mathit{occ}_i^c$ ranges in $0$ (no occlusion) to $1$ (all occluded). If $\bm{b}_i$ does not appear in this camera, we treat it is totally occluded. For multi-view, the calculation involves all cameras $C$ in a unique scene:
\begin{equation}
    \mathit{m\text{-}occ}_i = \operatorname{avg}(\{\mathit{occ}_i^c,\;c\in C\}),
\end{equation}
which is the averaged occlusion for a box over all views. Comparison of two metrics in \cref{Fig.OCC} shows the natural advantage of performing multi-view perception. Proportion of objects totally occluded under monocular view ($\mathit{occ} = 1$) is approximately $18\%$, while it decreases to $5\%$ in the case of multi-view.

\noindent{\textbf{Evaluation Protocol.}}
Since our task involves multi-view perception similar to that of nuScenes~\cite{nuScenes}, we follow the evaluation protocol outlined in the same work to assess the model performance. Specifically, the mAP is computed using matching thresholds of $\left\{0.5, 1, 2, 4\right\}$ to assess the detection performance. For true positives, we report average translation error, scale error and orientation error (ATE, ASE, AOE) for additional evaluation. The final nuScenes detection score (NDS) is calculated by a weighted average over these metrics. Please refer to their original paper for details.

\section{RoBEV for RoScenes}
In this section, we propose RoBEV, a novel method for effective 3D BEV detection on RoScenes. RoBEV is designed within the implicit BEV perception paradigm, which prioritizes efficiency in the context of a
large perception area.
To make precise 3D prediction, RoBEV relies on correctly assigning 2D perspective image features into 3D BEV features. This assignment process is largely affected by 3D position embedding attached on image features.
However, addressing the challenges mentioned  in RoScenes (Sec.~\ref{Sec.Intro}), particularly in our study, becomes challenging due to the varying sensor layouts across RoScenes, making the design of an effective 3D position embedding a complex task.
Therefore, it is imperative to explore an effective 3D position embedding 
in the 2D-3D feature assignment process.

\begin{figure}[t]
    \centering
    \begin{subfigure}[b]{0.58\linewidth}
    \centering
    \includegraphics[width=\linewidth]{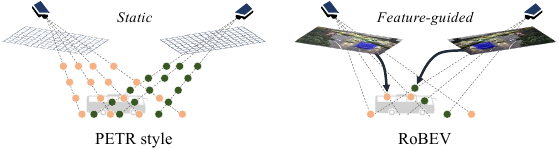}
    \caption{}
    \label{Fig.Frustum}
    \end{subfigure}
    \hfill
    \begin{subfigure}[b]{0.4\linewidth}
    \centering
    \includegraphics[width=\linewidth]{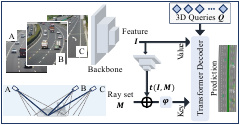}
    \caption{}
    \label{Fig.Framework}
    \end{subfigure}
    \caption{(a): Positional encoding of PETR and ours. (b): Framework of RoBEV.}
\end{figure}

We begin by outlining the fundamental principle of 2D-3D feature assignment, which serves as the central concept behind implicit BEV-based multi-view 3D object detection.
It learns a set of 3D position queries on BEV $\bm{q} \subseteq \mathbb{R}^D$ to aggregate 2D features from perspective image features $\bm{\mathcal{I}}\subseteq\mathbb{R}^{\left|C\right|\times H \times W \times D}$ via cross attention~\cite{Attention}, where $D$ is feature dimension, $\left|C\right|$ is number of cameras, $H, W$ are image feature size. The 2D-3D feature assignment
is heavily influenced by the cross-correlation between the learned BEV position queries and the image position embeddings.
Denoting the set of rays that traverse all the images as $\bm{\mathcal{M}}$, the cross-correlation can be formulated as:
\begin{equation}
    \langle\bm{q}, \phi\left(s\left(\bm{\mathcal{M}}\right)\right)\rangle
\end{equation}
where $\langle\cdot, \cdot\rangle$ is inner product, and $s(\cdot)$ samples a fixed number of 3D points that are uniformly distributed along the camera ray.
Here, $\phi(\cdot)$ is a learnable network to produce camera-specific spatial information by learning from these points, depicted in the left part of \cref{Fig.Frustum}.
However, it is important to note that the learning process of $\phi(\cdot)$ is constrained to the specific camera view, without taking into account the variations of different camera layouts.

Therefore, we
propose an enhanced feature-guided position embedding that leverages contextual information from $\bm{\mathcal{I}}$ to augment feature assignment process:
\begin{equation}
    \langle\bm{q}, \varphi\left(t\left(\bm{\mathcal{I}}, \bm{\mathcal{M}}\right)\right)\rangle
    \label{Eq.CrossAttn}
\end{equation}
where $t(\cdot)$ predicts a single 3D point for each pixel along the camera ray $\bm{\mathcal{M}}$, based on the image features $\bm{\mathcal{I}}$.
In this way, the learnable network $\varphi(\cdot)$ is able to represent camera layout-aware spatial information. Meanwhile, $\bm{q}$ is now obtained by transforming 3D queries with the same transformation $\varphi(\cdot)$ to ensure $\bm{q}$ and $t\left(\bm{\mathcal{I}}, \bm{\mathcal{M}}\right)$ are in same feature space. The framework is illustrated in \cref{Fig.Framework}.
The main difference of our feature assignment design consists of two perspectives. Firstly, it is constrained to learn from 3D image reference points in BEV perception cuboid other than using discretized camera frustum. This makes learning of $\varphi(\cdot)$ easier and gives a faster convergence. Secondly, the process is dynamic and feature-guided, making it easy to adapt to different sensor layouts. Later in next section, we would show its effectiveness in RoScenes.

\begin{table}[t]
\centering
\caption{Train, validation (easy, hard) and unseen test splits for benchmark. We report the number of clips and total images.}
\label{Tab.Split}

\resizebox{0.6\linewidth}{!}{
\begin{tabular}{@{}lrrrc@{}}
\toprule
Scene & \multicolumn{1}{c}{Train}& \multicolumn{1}{c}{~Easy}  & \multicolumn{1}{c}{~Hard} & \multicolumn{1}{c}{Test (Unseen)}  \\ \midrule
\#001 &$329 (138k)$ &$51 (21k)$ &$51 (21k)$ & \multirow{4}{*}{~~\#005 $\sim$ \#014~~} \\
\#002 &$534 (256k)$ &$83 (40k)$ &$83 (40k)$ &                   \\
\#003 &$306 (129k)$ &$48 (20k)$ &$48 (20k)$ &                   \\
\#004 &$534 (257k)$ &$80 (40k)$ &$80 (40k)$ &                   \\ \midrule
Sum  &~~$1,703 (779k)$&~~$265 (121k)$&~~$265 (121k)$& $637 (274k)$ \\ \bottomrule
\end{tabular}
}
\end{table}

\section{Benchmark}
We perform comprehensive study on BEV-based multi-view 3D object detection on RoScenes. A brief description of setup is given below.

\noindent\textbf{Competing methods.} In the benchmark, we adopt two groups of methods: 1) Explicit methods: BEVDet~\cite{BEVDet}, BEVDet4D~\cite{BEVDet4D}, SOLOFusion~\cite{SOLOFusion} and BEVFormer~\cite{BEVFormer}. 2) Implicit methods: DETR3D~\cite{DETR3D}, PETRv2~\cite{PETRv2}, StreamPETR~\cite{StreamPETR} and our RoBEV. The difference between two groups can be found in \cref{Sec.RelatedWorks}.

\noindent\textbf{Implementation details.} To ensure a fair comparison, we evaluate all methods under MMDetection3D framework, based on PyTorch~\cite{MMDet3D,PyTorch}. The input image size is $576 \times 1024$. The backbone for image feature extraction is VoVNetV2-99~\cite{V299} (pretrained on nuScenes). The remaining part in each method follows their default setting. All methods are trained for $12$ epochs on a machine with $8$ NVIDIA Tesla V100 GPUs, batch size $= 8$, and are with a cosine annealing learning rate schedule where initial learning rate is $2\times{10}^{-4}$. No extra training/inference strategies are employed, \eg, CBGS~\cite{CBGS} or test-time augmentation.

To validate model performance under different conditions, the dataset is split based on cip-level multi-view occlusion, which is the average $m\text{-}occ$ for all vehicles in an entire clip. We choose \#001$\sim$\#004 for train and validation, while using \#005$\sim$\#014 for test. For the former, we sort all clips via clip-level multi-view occlusion and choose bottom/top $10\%$ as easy/hard val set (threshold is $<0.23$ and $>0.48$), and remaining data becomes train set. The train, validation and test sets are shown in \cref{Tab.Split}.

\begin{table}[t]
    \centering
    \caption{Performance comparison of BEV methods on RoScenes dataset.}
    \label{Tab.Results}
    \resizebox{\linewidth}{!}{
    \begin{tabular}{@{}lrrrrrrrrrrc@{}}
    \toprule
     \multirow{2}{*}{Method}    & \multicolumn{5}{c}{Easy} & \multicolumn{5}{c}{Hard} & \multirow{2}{*}{\begin{tabular}[c]{@{}c@{}}\textit{Avg.}\\ NDS\end{tabular}} \\ \cmidrule(lr){2-6}\cmidrule(lr){7-11}
        & NDS & mAP & mATE & mASE & mAOE & NDS & mAP & mATE & mASE & mAOE & \\ \midrule
     BEVDet~\cite{BEVDet}    &$0.506$&$0.299$&$0.742$&$0.079$&$0.042$&$0.445$&$0.184$&$0.754$&$0.087$&$0.043$&$0.476$   \\
     BEVDet4D~\cite{BEVDet4D}   &$0.428$&$0.200$&$0.896$&$0.094$&$0.041$&$0.393$&$0.139$&$0.922$&$0.099$&$0.038$&$0.411$  \\
     SOLOFusion~\cite{SOLOFusion}  &$0.308$&$0.129$&$0.878$&$0.144$&$0.517$&$0.202$&$0.066$&$0.844$&$0.144$&$1.000$&$0.255$  \\
     BEVFormer~\cite{BEVFormer} &$0.693$&$0.609$&$0.560$&$0.078$&$0.030$ &$0.597$&$0.433$&$0.600$&$0.090$&$0.029$&$0.645$ \\\midrule
     DETR3D~\cite{DETR3D}    &$0.722$&$0.644$&$0.501$&$0.067$&$0.031$&$0.633$&$0.471$&$0.508$&$0.080$&$0.028$&$0.678$\\
     PETRv2~\cite{PETRv2}      & $0.674$ & $0.587$ & $0.590$ & $0.090$ & $0.032$ & $0.580$ & $0.414$& $0.633$ & $0.100$ & $0.029$  &$0.627$\\
     StreamPETR~\cite{StreamPETR}    &$0.619$ & $0.513$ & $0.690$ & $0.102$ & $0.032$ &$0.496$ & $0.284$ & $0.739$ & $0.107$ & $0.031$&$0.558$ \\
     \textbf{RoBEV (Ours)} &$\mathbf{0.753}$ & $\mathbf{0.684}$ & $0.442$ & $0.058$ & $0.031$ &$\mathbf{0.672}$ & $\mathbf{0.524}$ & $0.438$ & $0.077$ & $0.027$&$\mathbf{0.713}$ \\
    \bottomrule
    \end{tabular}
    }

    \end{table}

\subsection{Result on Validation Set}
We use all scenes' train set to train models and report validation set's results in \cref{Tab.Results}. In general, implicit methods achieve $19.7\%$ higher performance in average than explicit ones. The performance gain comes from discard of BEV feature map. Explicit methods project each 2D image feature to dense BEV feature by weights along camera ray. RoScenes' perception area makes the process involve a large set of weights which may hard to learn and leads to low performance. BEVDet4D and SOLOFusion are inferior to the baseline method BEVDet, the problem may stem from absence of depth-map supervision and CBGS training. For implicit ones, the PETR series achieve a much worse performance than DETR3D because of the static position embedding as aforementioned. Our RoBEV fixes this issue and obtains significant performance gain. It is worth noting RoScenes has a large variance in velocity as shown in \cref{Fig.Stat} and most methods can not achieve reasonable accuracy for velocity prediction (mAVE = 1). We will explore this in future study.

\begin{table}[t]
\centering
\caption{\textbf{Left}: NDS comparison of DETR3D, PETRv2 and RoBEV with different backbones. ResNets and V2-99$^\ddagger$ are pretrained on ImageNet. \textbf{Right}: Transferability validation between a single scene \#001 and all scenes.}
\label{Tab.Backbone}
 \aboverulesep=0ex
 \belowrulesep=0ex
 \renewcommand{\arraystretch}{1.1}
\resizebox{\linewidth}{!}{
\begin{tabular}{lcccccccc||cccccccc}\toprule
\multirow{2}{*}{Method} & \multicolumn{2}{c}{Res-50} & \multicolumn{2}{c}{Res-101}  & \multicolumn{2}{c}{V2-99$^\ddagger$}  & \multicolumn{2}{c||}{V2-99} & \multicolumn{2}{|c}{Single$\rightarrow$Single}&\multicolumn{2}{c}{Single$\rightarrow$All}&\multicolumn{2}{c}{All$\rightarrow$Single}&\multicolumn{2}{c}{All$\rightarrow$All} \\ \cmidrule(lr){2-3}\cmidrule(lr){4-5}\cmidrule(lr){6-7}\cmidrule(lr){8-9}\cmidrule(lr){10-11}\cmidrule(lr){12-13}\cmidrule(lr){14-15}\cmidrule(lr){16-17}
                        &  Easy & Hard & Easy & Hard & Easy & Hard & Easy & Hard &  Easy  & Hard    &  Easy  & Hard&  Easy  & Hard    &  Easy  & Hard \\\midrule
DETR3D   & $0.707$ & $0.594$ & $0.709$ & $0.601$ & $0.721$ & $0.654$ & $0.722$ & $0.633$ & $0.660$ & $0.545$ & $0.382$ & $0.375$ &$0.701$ & $0.614$ & $0.722$ & $0.633$     \\
PETRv2   & $0.545$ & $0.407$ & $0.574$ & $0.435$ & $0.665$ & $0.577$ & $0.674$ & $0.580$ &$0.649$ & $0.512$ & $0.376$ &   $0.359$&$0.636$ & $0.563$ & $0.674$ & $0.580$     \\
RoBEV (Ours) & $0.726$ & $0.633$ & $0.729$ & $0.638$ & $0.742$ & $0.671$ & $0.753$ & $0.672$ &$0.683$ & $0.571$ & $0.396$ &     $0.387$&$0.720$ & $0.631$ & $0.753$ & $0.672$ \\\bottomrule
\end{tabular}
}
\end{table}

We also compare computational costs of these methods in \cref{Fig.Performance}. Due to extra operations on the huge BEV feature map, explicit methods have $1.75\times$ training time and $1.50\times$ inference memory cost than implicits. In summary, we recommend future studies of BEV-based methods in RoScenes to focus on implicit or more efficient paradigms.




\begin{table}[t]
    \centering
    \caption{Ablation study of RoBEV in terms of positional encoding and backbone.}
    \label{Tab.Ablation}
    \resizebox{0.8\linewidth}{!}{
    \begin{tabular}{@{}lccccccccccc@{}}
    \toprule
     \multirow{2}{*}{Variants}    & \multicolumn{5}{c}{Easy} & \multicolumn{5}{c}{Hard} & \multirow{2}{*}{\begin{tabular}[c]{@{}c@{}}\textit{Avg.}\\ NDS\end{tabular}} \\ \cmidrule(lr){2-6}\cmidrule(lr){7-11}
        & NDS & mAP & mATE & mASE & mAOE & NDS & mAP & mATE & mASE & mAOE & \\ \midrule
     Sinusoidal $\varphi$    &$0.664$ & $0.573$ & $0.603$ & $0.098$ & $0.032$ & $0.551$ & $0.361$ & $0.638$ & $0.017$ & $0.030$ & $0.608$  \\
     Separate $\varphi$   &$0.721$ & $0.637$ & $0.490$ & $0.065$ & $0.031$ & $0.649$ & $0.496$ & $0.489$ & $0.077$ & $0.029$ & $0.685$  \\\midrule
     RoBEV &$0.753$ & $0.684$ & $0.442$ & $0.058$ & $0.031$ &$0.672$ & $0.524$ & $0.438$ & $0.077$ & $0.027$&$0.713$ \\
    Swin-B & $0.729$ & $0.651$ & $0.466$ & $0.080$ & $0.031$ & $0.664$ & $0.515$ & $0.449$ & $0.080$ & $0.027$ & $0.697$ \\
    ViT-L & $0.759$ & $0.679$ & $0.434$ & $0.056$ & $0.030$ & $0.679$ & $0.535$ & $0.427$ & $0.074$ & $0.030$ & $0.719$ \\
    Intern-XL & $0.762$ & $0.692$ & $0.421$ & $0.056$ & $0.028$ & $0.676$ & $0.530$ & $0.432$ & $0.075$ & $0.028$ & $0.719$\\
    \bottomrule
    \end{tabular}
    }
    \end{table}

\subsection{Ablation Study on Validation Set}
\noindent\textbf{Impact of positional encoding.} We design two variants of RoBEV for ablation study \wrt positional encoding $\varphi$, which is shown in \cref{Tab.Ablation}. The first one, \enquote{sinusoidal $\varphi$} uses the parameter-free sinusoidal positional encoding~\cite{Attention} to transform points (\cref{Eq.CrossAttn}) to $D$-dim embedding. The second, \enquote{Separate $\varphi$} uses two independent learnable networks to transform queries and points other than a shared network. Both variants achieve lower performance in validation set.

\noindent\textbf{Impact of backbone.} The backbone is replaced with ResNet~\cite{ResNet} (with deformable convolution~\cite{DeformConv}) to show the impact of different backbone architectures. Moreover, we also train a model with VoVNetV2-99 pretrained on ImageNet~\cite{ImageNet} to evaluate the impact of pretraining data. In \cref{Tab.Backbone} left, models with ResNets have a $8.1\%$ drop against VoVNet, showing the effectiveness of latter's design in 3D detection task. When using ImageNet pretrained VoVNet as backbone, there is $0.6\%$ drop compared to nuScenes pretrained model. Therefore, pretraining images that are in similar domain, \eg, vehicle-side, can enhance performance. We further validate the performance with Swin-B ($88M$ parameters)~\cite{Swin}, ViT-L~\cite{ViT} from SAM~\cite{SAM} ($308M$ parameters) and InternImage-XL~\cite{InterImage} ($387M$ parameters). We empirically find these models need more iterations to achieve similar loss compared to above backbones. Therefore, we extend training epoch to $48$ and report result in \cref{Tab.Ablation}. However, we could not observe noticeable performance gain. It suggests current backbone has enough capacity for perception.

\begin{table}[t]
    \centering
    \caption{Performance comparison on zero-shot test set.}
    \label{Tab.ZeroShot}
    \resizebox{0.5\linewidth}{!}{
    \begin{tabular}{@{}lrrrrr@{}}
    \toprule
     \multicolumn{1}{c}{Method}   & NDS & mAP & mATE & mASE & mAOE \\ \midrule
     BEVDet~\cite{BEVDet} & $0.308$ & $0.019$ & $0.981$ & $0.141$ & $0.090$     \\
     BEVDet4D~\cite{BEVDet4D} & $0.298$ & $0.016$ & $1.000$ & $0.154$ & $0.105$   \\
     SOLOFusion~\cite{SOLOFusion}  & $0.304$ & $0.016$ & $1.000$ & $0.133$ & $0.089$  \\
     BEVFormer~\cite{BEVFormer} & $0.308$ & $0.022$ & $1.000$ & $0.138$ & $0.082$  \\\midrule
     DETR3D~\cite{DETR3D} & $0.304$ & $0.024$ & $1.000$ & $0.143$ & $0.104$    \\
     PETRv2~\cite{PETRv2}   & $0.288$ & $0.011$ & $1.000$ & $0.174$ & $0.129$    \\
     StreamPETR~\cite{StreamPETR}  & $0.287$ & $0.009$ & $1.000$ & $0.175$ & $0.130$  \\
    RoBEV (Ours) & $0.308$ & $0.024$ & $1.000$ & $0.139$ & $0.084$  \\
    \bottomrule
    \end{tabular}
    }
    \end{table}

\begin{figure}[t]
    \centering
    \begin{subfigure}[b]{0.7\linewidth}
    \includegraphics[width=\linewidth]{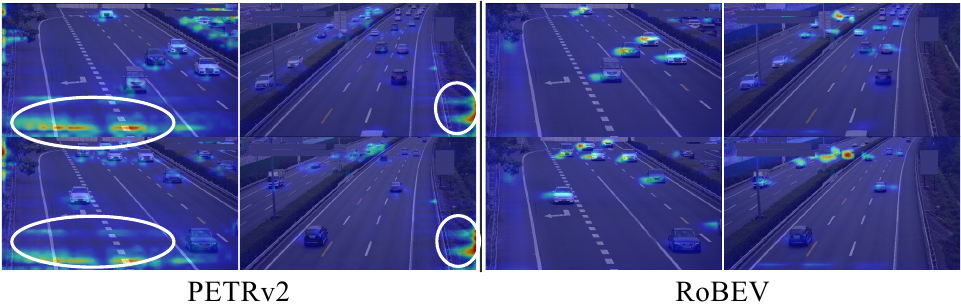}
    \caption{}\label{Fig.Visualiztion}
    \end{subfigure}
    \hfill
    \begin{subfigure}[b]{0.28\linewidth}
    \includegraphics[width=\linewidth]{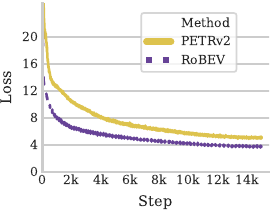}
    \caption{}\label{Fig.LossCurve}
    \end{subfigure}
    \caption{(a): Attention heatmap visualization. PETRv2 has static artifacts in heatmap across scenes and cameras. (b): Convergence curves of PETRv2 and RoBEV.}
\end{figure}

\noindent\textbf{Impact of cross-scene training.} Training across scenes has strong impacts for model performance. In \cref{Tab.Backbone} right, we report NDS with different train/val splits. When we use a single scene \#001 for training, both methods fit to this static layout well, and achieve reasonable performance. However, they show $17.9\%$ performance drop in average when validated on all scenes. Nevertheless, single-scene trained RoBEV outperforms the competitor by $2.4\%$ in this case, indicating our feature-guided position embedding is better to adapt to novel layouts. The performance gap between PETRv2 and ours is enlarged to $8.6\%$ when performing a full data train and validation. In this setting, PETRv2's static position embedding can not handle variation of sensor layouts. We also show loss and convergence comparison in \cref{Fig.LossCurve} which can be an evidence. The full data training also enhances model performance when tested on the aforementioned single scene. Especially, our RoBEV has a $4.9\%$ higher NDS than single scene trained. This indicates training with a mixed sensor layout, large sensor variety and more data benefits for model capacity.

\noindent\textbf{Attention Visualization.} We visualize cross attention heatmap between query and the combination of image features and position embedding for both PETRv2 and our RoBEV in \cref{Fig.Visualiztion}. PETRv2's attention map exhibits static artifacts
across cameras and scenes, indicating a flaw in its use of static position embedding. In contrast, our RoBEV correctly locates attention on target vehicles and is dynamic based on image content, showing the effectiveness of our proposed feature-guided position embedding.

\subsection{Result on Zero-Shot Test Set}
We emphasize the key difference of RoScenes benchmark is the test on unseen zero-shot scenes. The road and camera layouts are significantly different and strictly unknown compared to training data. The test result is placed in \cref{Tab.ZeroShot}. We observe a significant performance drop for all tested methods, that none of them achieves $>0.35$ NDS and $>0.1$ mAP. It indicates poor transferability of these BEV models. Therefore, we encourage the community to take study on RoScenes for BEV perception under various layouts, which is under explored but extremely important for applying BEV paradigm in real-word scenarios.

Besides, robustness evaluation for BEV models, impact of number of BEV queries, and monocular 3D detection benchmark are in appendix.

\section{Conclusion}
In this paper, we introduce the largest multi-view roadside perception dataset, RoScenes. It has a large perception range of $64,000 m^2$, a full scene coverage for various conditions, and a crowded traffic with a total of $21.13M$ annotated vehicles. A novel BEV-to-3D data pipeline is provided to ensure annotation efficiency.
Moreover, we conduct benchmark on RoScenes and propose RoBEV to handle the variation of sensor layouts.
Our method significantly outperforms state-of-the-arts by a large margin on validation set. The dataset, algorithmic baselines and a toolkit will be made available.

\noindent\textbf{Limitation and Future Work.}
The main limitation of our dataset is the lack of task diversity in real-world applications. In future work, we will extend RoScenes to support single/cross-scene long-term tracking, prediction and multimodal fusion tasks. Meanwhile, due to inadequacy of UAV pipeline, data location and privacy issues, other weather conditions and roadside scenes are currently not included. We will extend to include tunnels, intersections, and foggy, rainy clips in the future. Moreover, both ours and existing BEV algorithms have unsatisfied performance when adapting to unseen scenes. The attempts to explore more effective and robust roadside BEV approaches for transferability are needed.

%
%
\bibliographystyle{splncs04}
\bibliography{main}

\appendix

\renewcommand{\thefigure}{\alph{figure}}
\renewcommand{\thetable}{\alph{table}}


\clearpage
\section*{Appendix}

\section{Additional Dataset Analysis}
\subsection{Scene Samples}
\begin{figure}[htp!]
         \vspace{2em}
    \begin{subfigure}[b]{\linewidth}
    \centering
    \includegraphics[width=\linewidth]{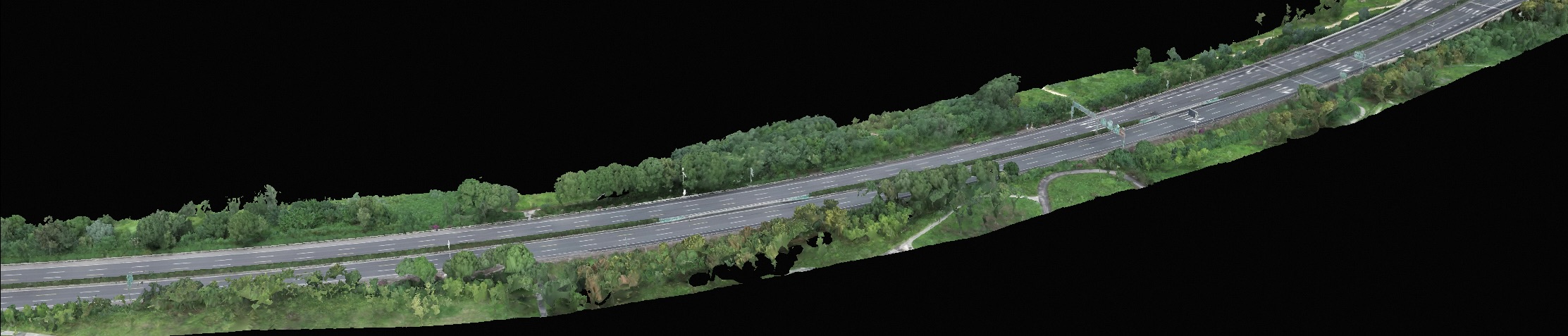}
    \caption{The 3D reconstruction of \#001.}
         \vspace{2em}
    \end{subfigure}
    \begin{subfigure}[b]{\linewidth}
    \centering
    \includegraphics[width=\linewidth]{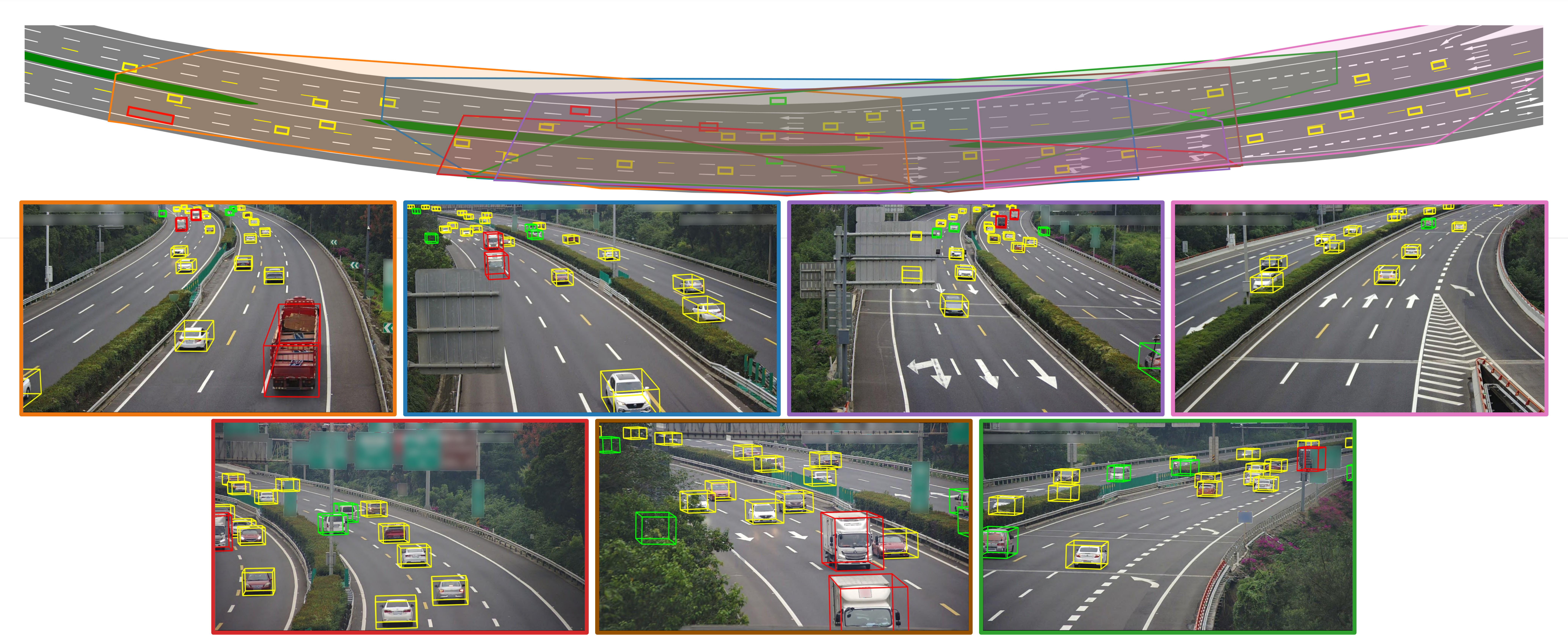}
    \caption{The Day-Normal sample of \#001.}
         \vspace{2em}
    \end{subfigure}
    \begin{subfigure}[b]{\linewidth}
    \centering
    \includegraphics[width=\linewidth]{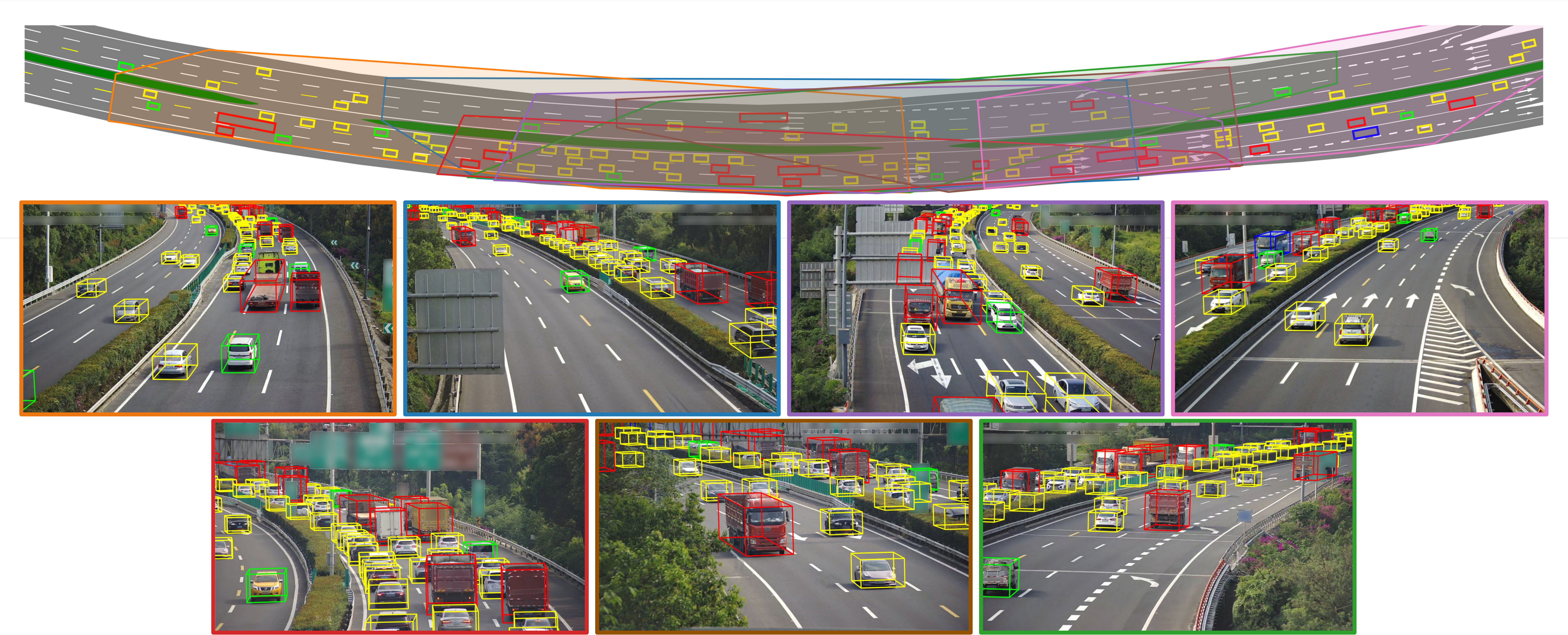}
    \caption{The Day-Heavy sample of \#001.}
    \end{subfigure}
    \caption{Sample visualization of \#001.}
    \label{Fig.Vis1}
\end{figure}

\begin{figure}[H]
         \vspace{2em}
    \begin{subfigure}[b]{\linewidth}
    \centering
    \includegraphics[width=\linewidth]{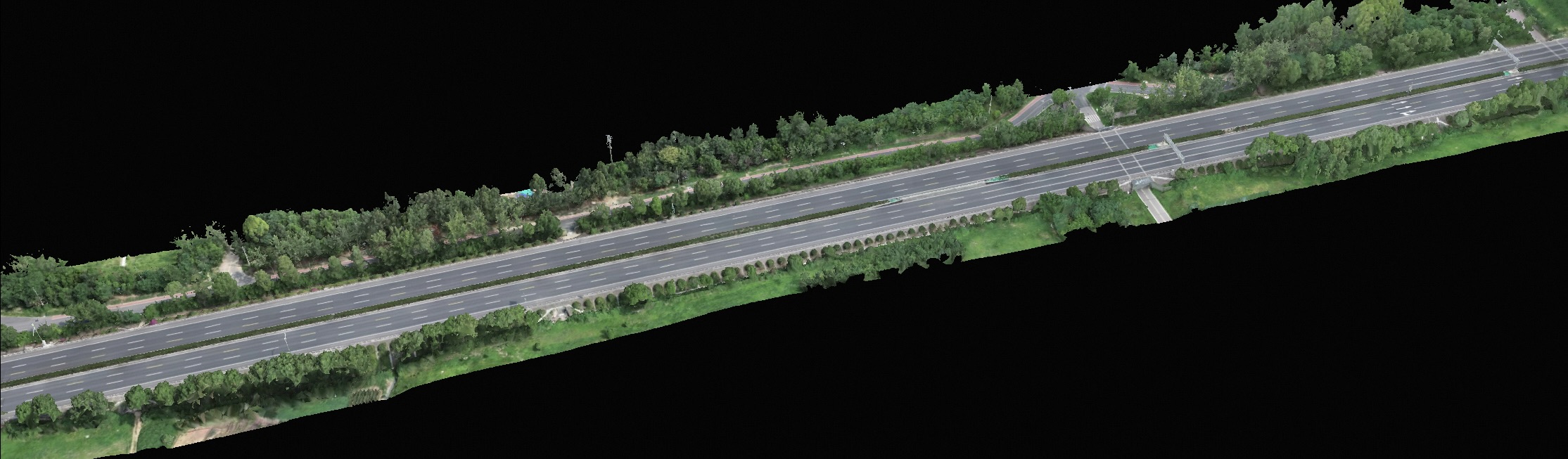}
    \caption{The 3D reconstruction of \#002.}
         \vspace{2em}
    \end{subfigure}
    \begin{subfigure}[b]{\linewidth}
    \centering
    \includegraphics[width=\linewidth]{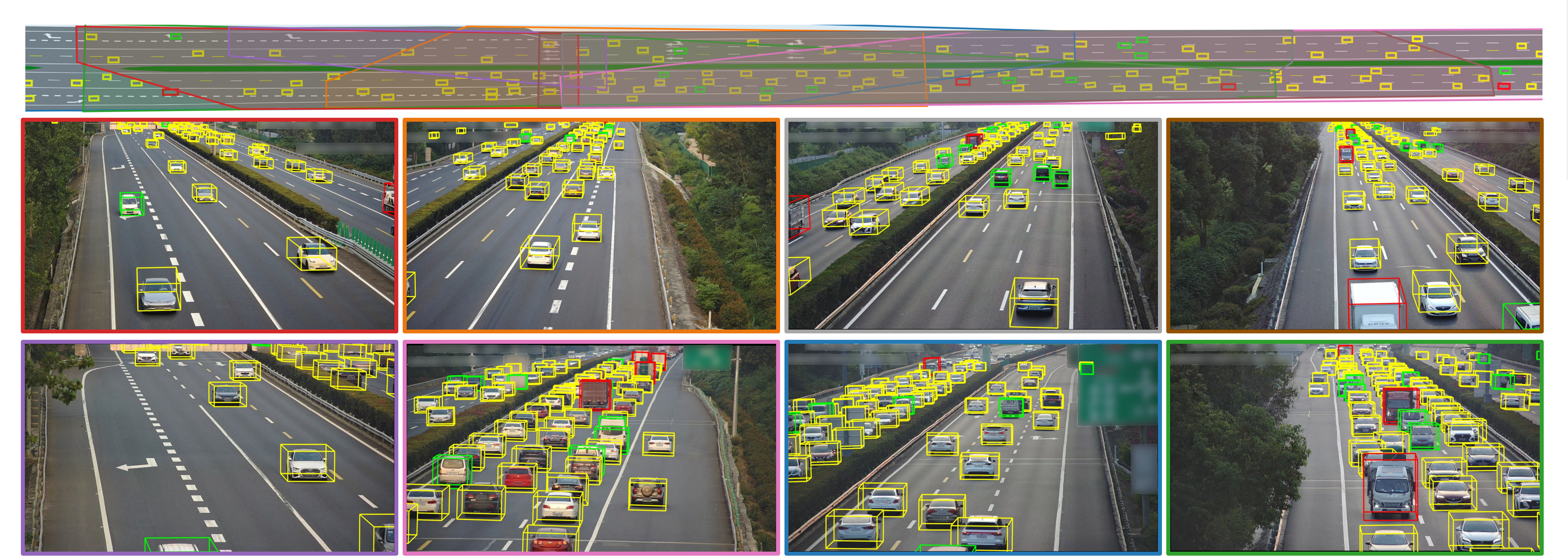}
    \caption{The Day-Normal sample of \#002.}
         \vspace{2em}
    \end{subfigure}
    \begin{subfigure}[b]{\linewidth}
    \centering
    \includegraphics[width=\linewidth]{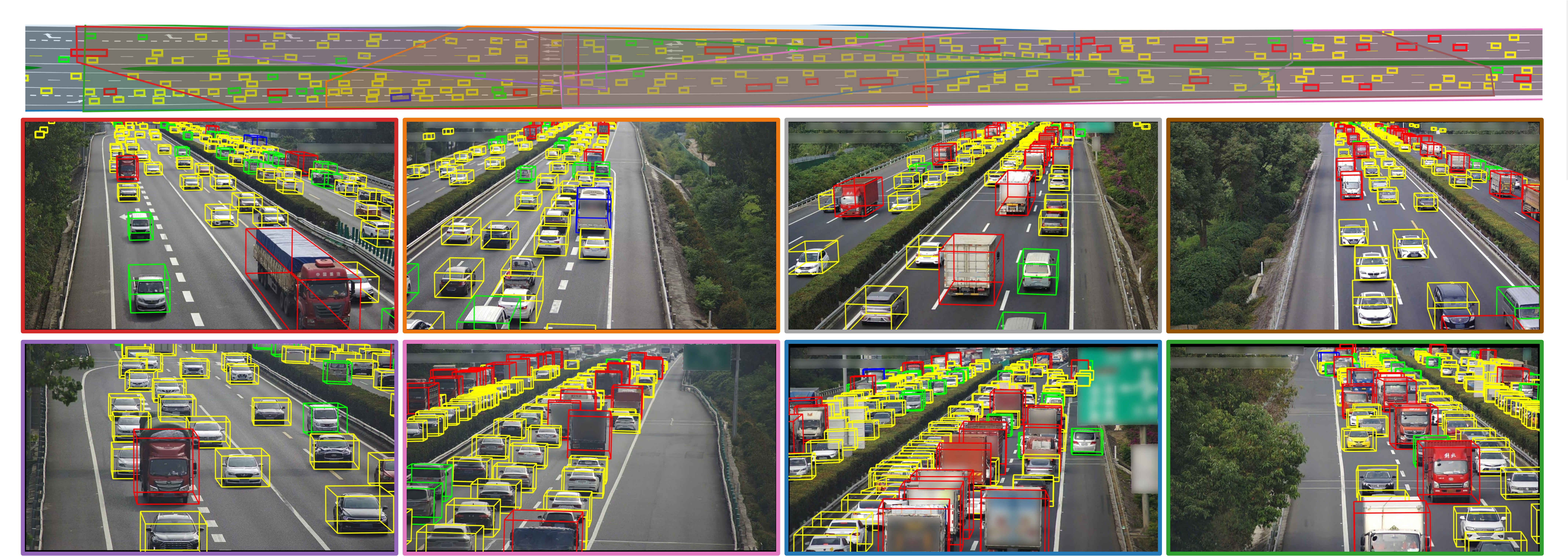}
    \caption{The Day-Heavy sample of \#002.}
    \end{subfigure}
    \caption{Sample visualization of \#002.}
    \label{Fig.Vis2}
\end{figure}

\begin{figure}[H]
         \vspace{2em}
    \begin{subfigure}[b]{\linewidth}
    \centering
    \includegraphics[width=\linewidth]{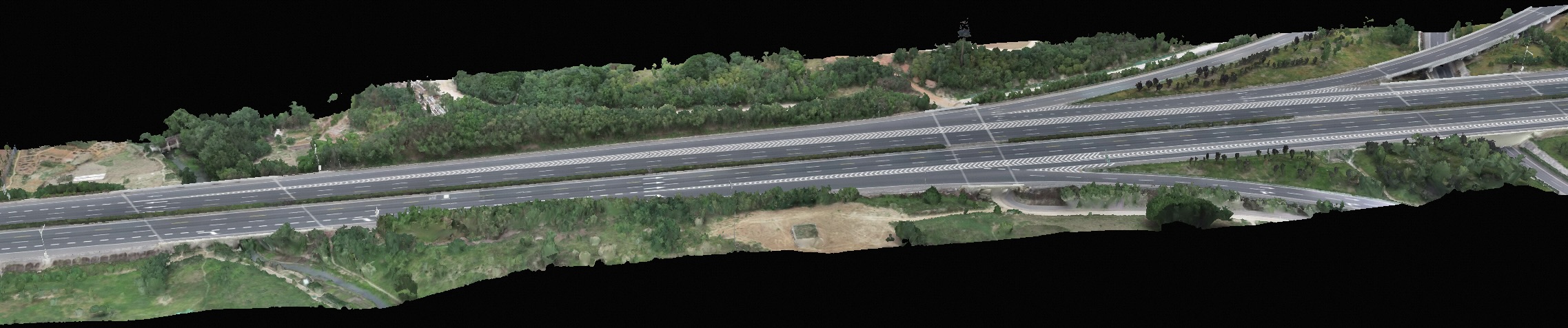}
    \caption{The 3D reconstruction of \#003.}
         \vspace{2em}
    \end{subfigure}
    \begin{subfigure}[b]{\linewidth}
    \centering
    \includegraphics[width=\linewidth]{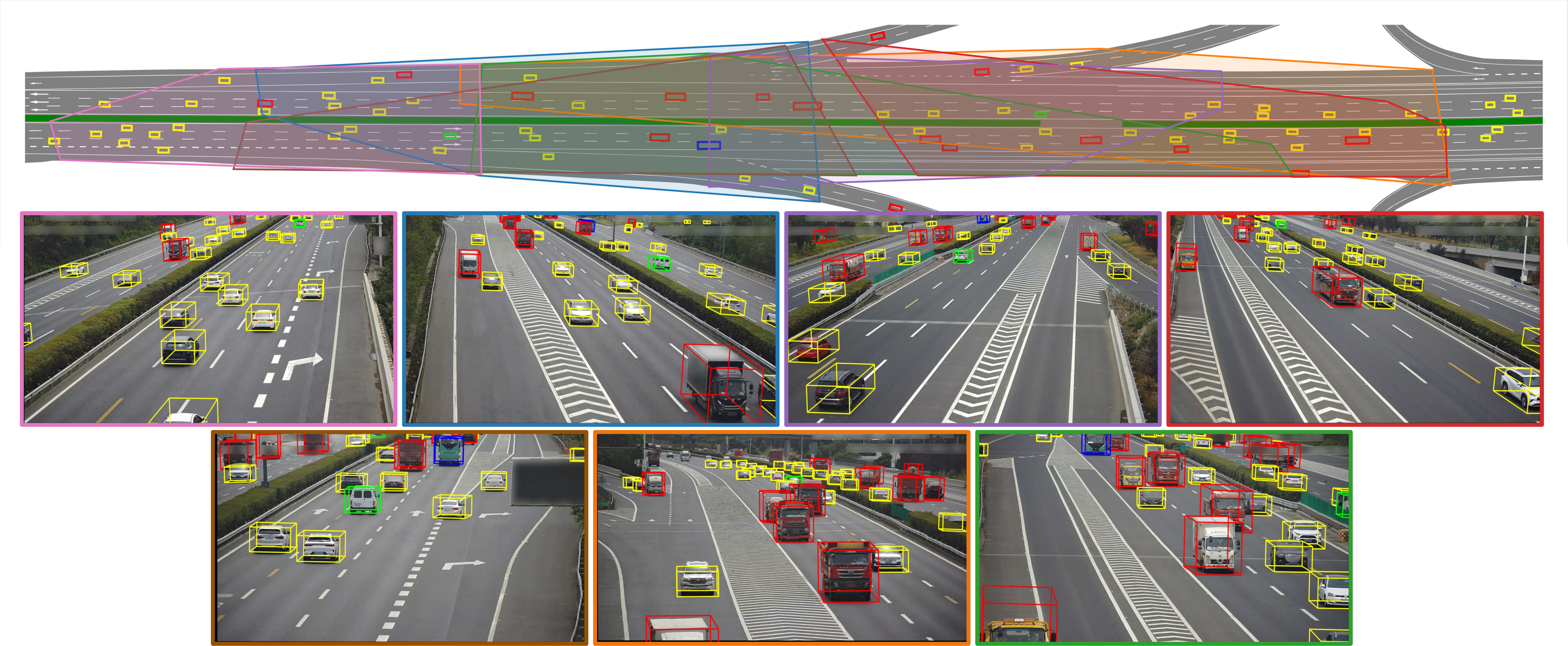}
    \caption{The Day-Normal sample of \#003.}
         \vspace{2em}
    \end{subfigure}
    \begin{subfigure}[b]{\linewidth}
    \centering
    \includegraphics[width=\linewidth]{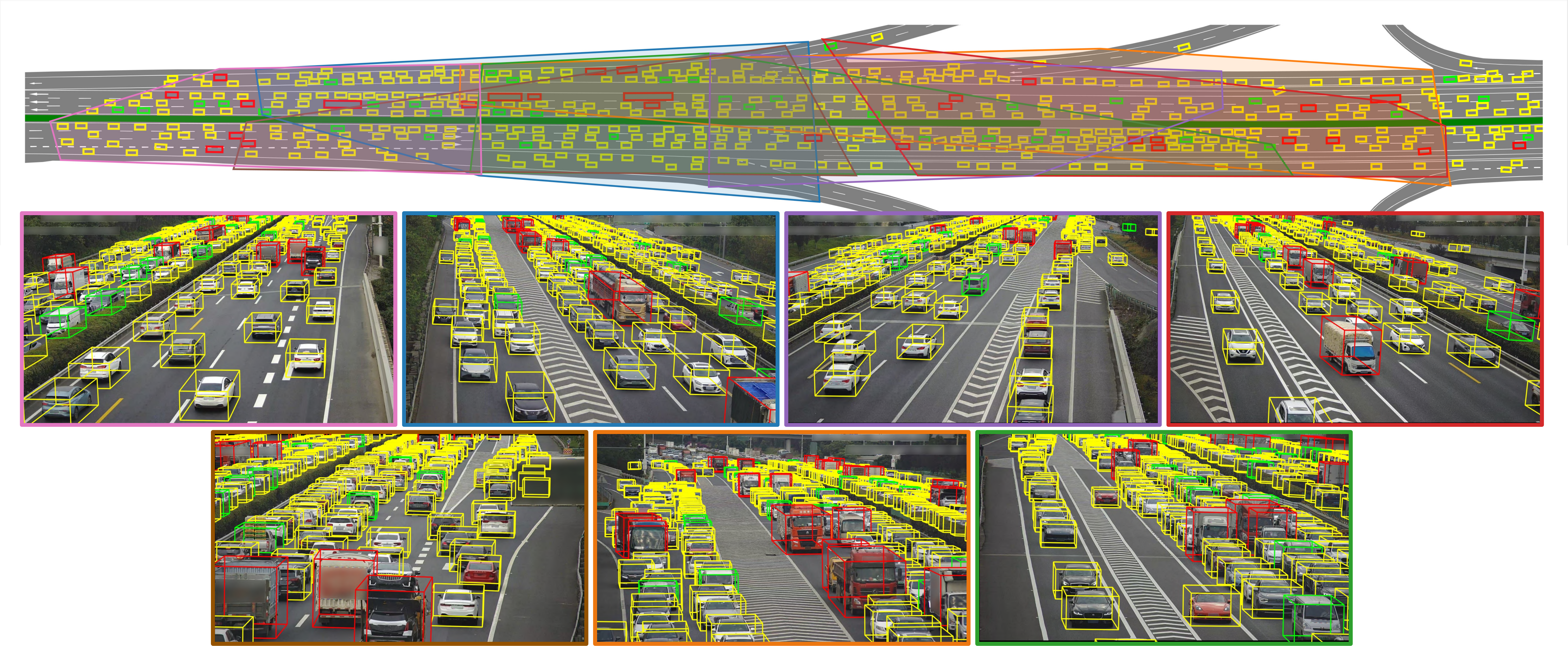}
    \caption{The Day-Heavy sample of \#003.}
    \end{subfigure}
    \caption{Sample visualization of \#003.}
    \label{Fig.Vis3}
\end{figure}

\begin{figure}[htp!]
         \vspace{2em}
    \begin{subfigure}[b]{\linewidth}
    \centering
    \includegraphics[width=\linewidth]{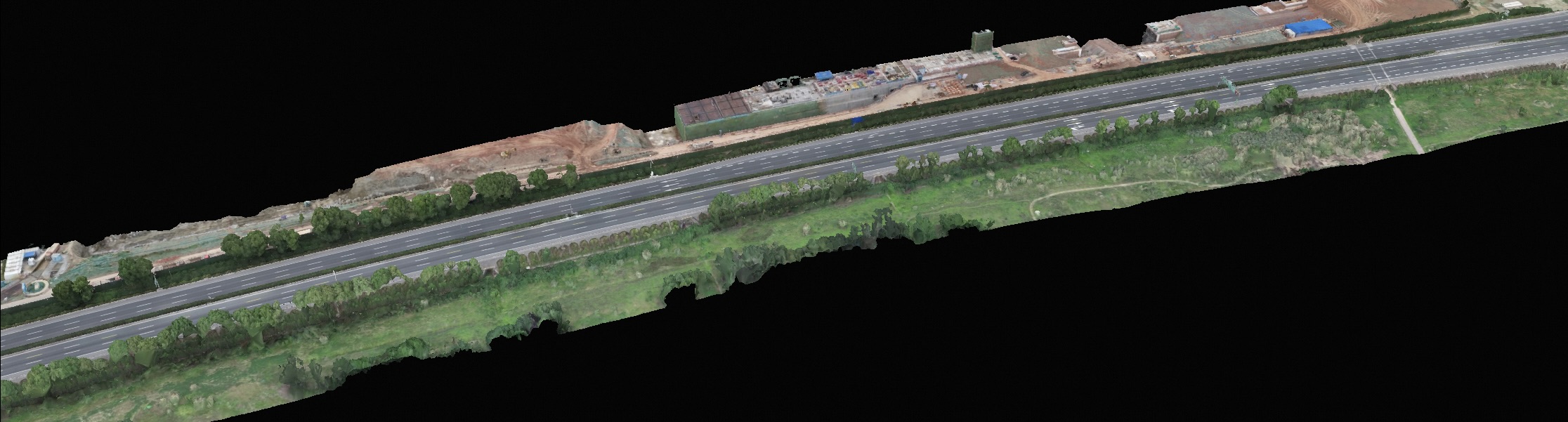}
    \caption{The 3D reconstruction of \#004.}
         \vspace{2em}
    \end{subfigure}
    \begin{subfigure}[b]{\linewidth}
    \centering
    \includegraphics[width=\linewidth]{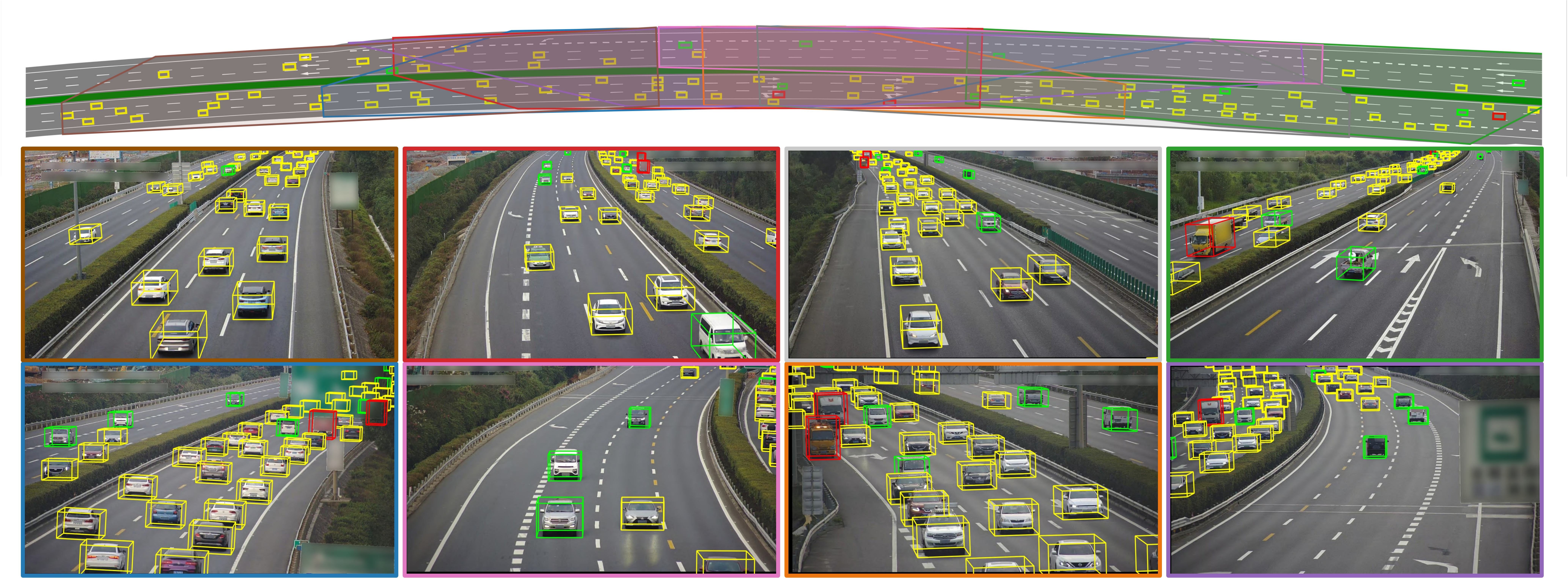}
    \caption{The Day-Normal sample of \#004.}
         \vspace{2em}
    \end{subfigure}
    \begin{subfigure}[b]{\linewidth}
    \centering
    \includegraphics[width=\linewidth]{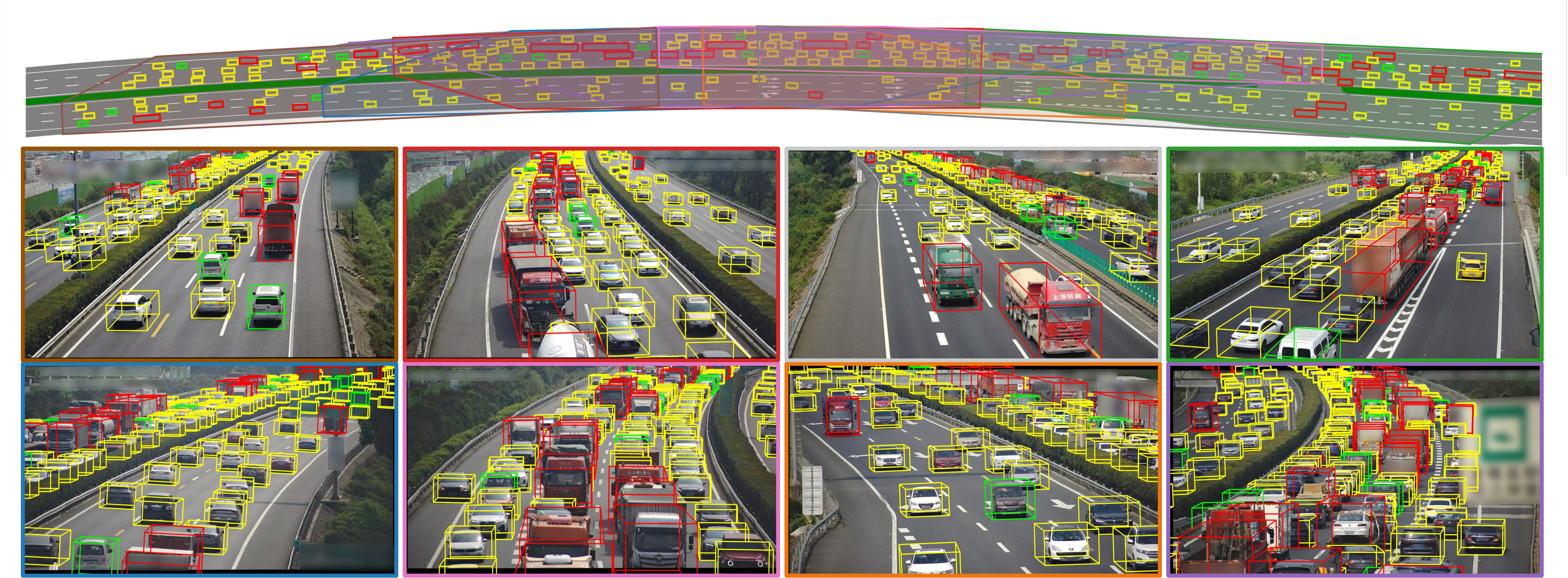}
    \caption{The Day-Heavy sample of \#004.}
    \end{subfigure}
    \caption{Sample visualization of \#004.}
    \label{Fig.Vis4}
\end{figure}

\begin{figure}[t]
    \centering
    \captionbox{The Night-Normal sample of \#004.\label{Fig.Vis5} \vspace{2em}}{
    \begin{subfigure}{\linewidth}
\centering
    \includegraphics[width=\linewidth]{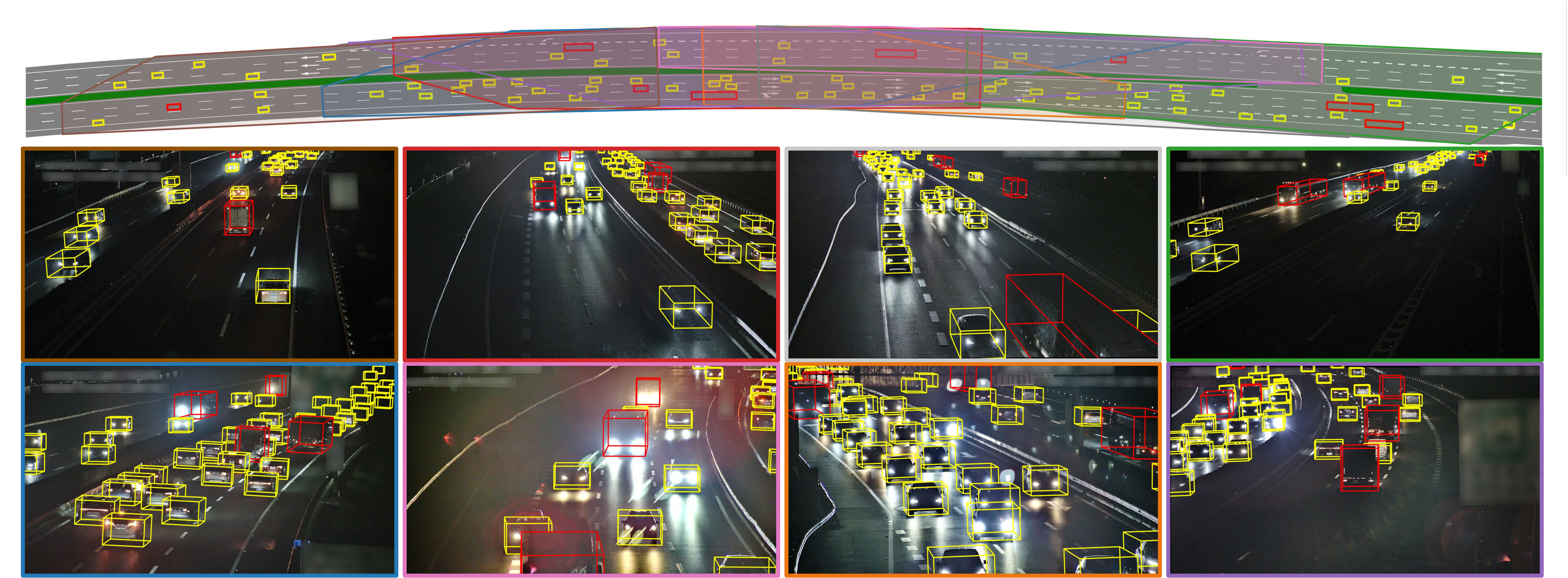}
    \end{subfigure}
    }
\begin{subfigure}{\linewidth}
\captionbox{\label{Fig.Orientation}}{
    \begin{subfigure}[b]{0.55\linewidth}
    \centering
    \includegraphics[width=\linewidth]{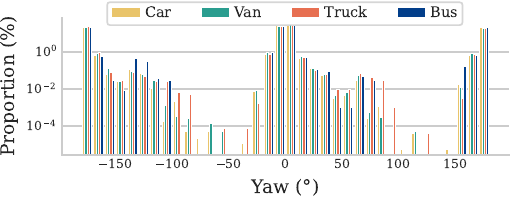}
    \end{subfigure}
    }
\hfill
    \captionbox{\label{Tab.Density}}{
    \begin{subtable}[b]{0.4\linewidth}
        \centering
        \resizebox{\linewidth}{!}{
    \begin{tabular}{@{}lrr@{}}\toprule
    Dataset &       ~~~~~~~~Small & ~~~~~~~~~~Big  \\\midrule
    KITTI~\cite{KITTI}   &  $4.2$ & $0.2$               \\
    ApolloScapes~\cite{ApolloScape} & $11.6$ & $0.0$       \\
    Argoverse~\cite{Argoverse}  &  $4.1$ & $0.3$             \\
    nuScenes~\cite{nuScenes}  &  $3.0$ & $0.6$               \\
    Waymo~\cite{Waymo}      &   $3.2$ & $0.0$            \\
    Rope3D~\cite{Rope3D}     &  $14.0$ & $0.6$         \\
    RoScenes (Ours) & $\mathbf{63.2}$ & $\mathbf{8.1}$            \\\bottomrule
    \end{tabular}}
    \end{subtable}
    }
\end{subfigure}
    \caption{More Statistics. (a): The statistic of Vehicle orientation. (b): The comparison of annotation density (\#box / image). \enquote{Car} and \enquote{Van} are in column of \enquote{small}. \enquote{Truck} and \enquote{Bus} are in column of \enquote{big}.}
\end{figure}

We provide the visualization of 3D reconstructions, BEV 2D annotations and perspective views for \#001, \#002, \#003, \#004 scenes with different traffic conditions in \cref{Fig.Vis1,Fig.Vis2,Fig.Vis3,Fig.Vis4,Fig.Vis5}. The high quality annotations verify the effectiveness of our data pipeline in building RoScenes.

\begin{figure}[H]
    \centering
    \begin{subfigure}[b]{0.49\linewidth}
        \centering
        \includegraphics[width=\linewidth]{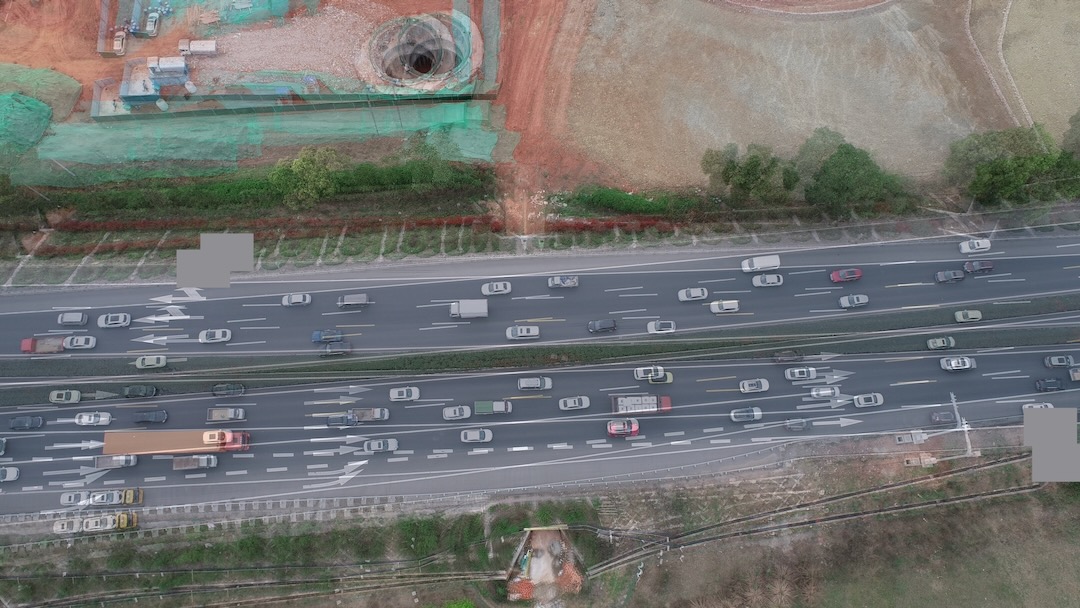}
        \caption{The overlay of two captured UAV images before registration.}
        \label{Fig.UAVPTZA}
    \end{subfigure}
    \begin{subfigure}[b]{0.49\linewidth}
        \centering
        \includegraphics[width=\linewidth]{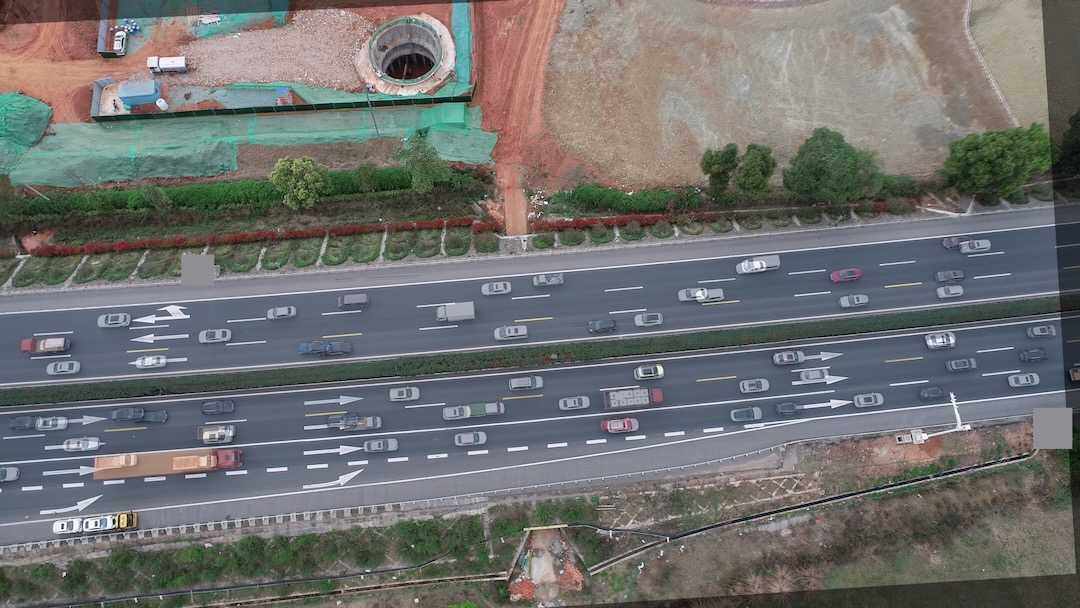}
        \caption{The overlay of two captured UAV images after registration.}
        \label{Fig.UAVPTZB}
    \end{subfigure}
    \caption{UAV image registration.}
    \label{Fig.UAVPTZ}
    \vspace{-4em}
\end{figure}

\begin{figure}[htp]
    \centering
    \includegraphics[width=0.7\linewidth]{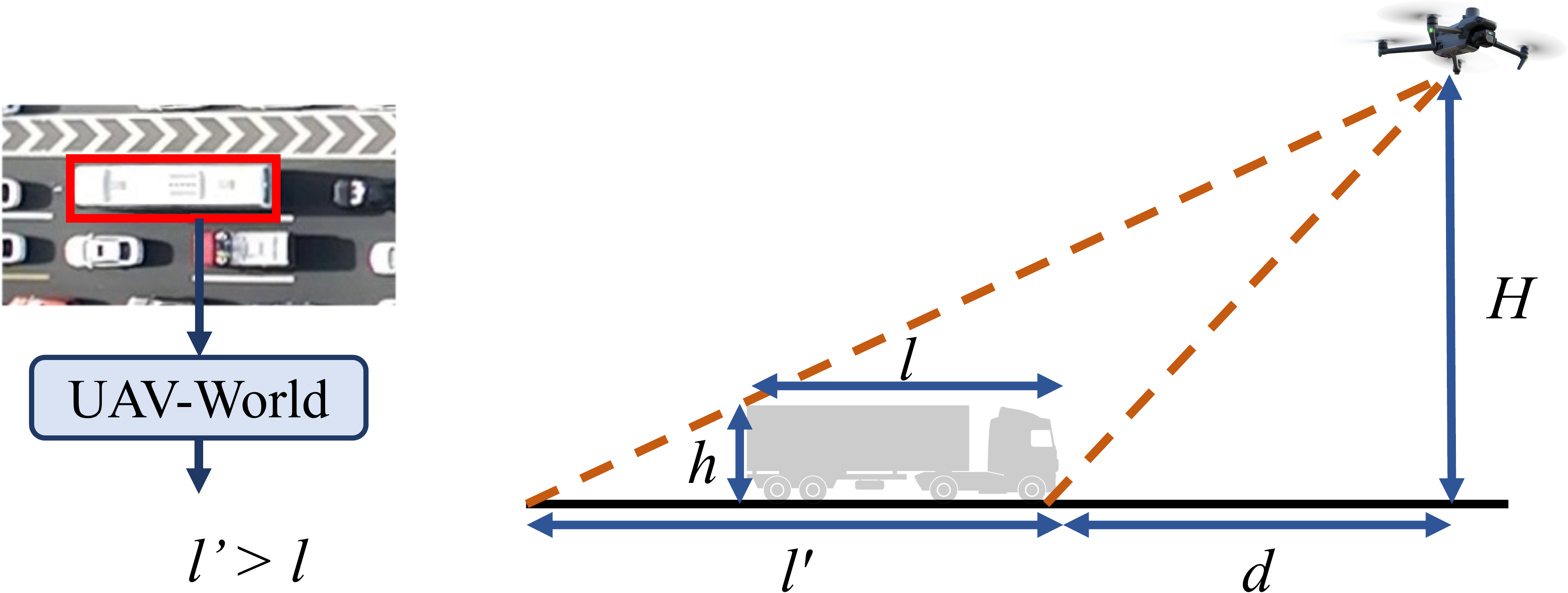}
    \caption{Length refinement for BEV 2D box.}
    \label{Fig.Perspective}
    \vspace{-4em}
\end{figure}

\subsection{Data Collection}
\label{Sec.Refine}
In this section, we would give a further discussion about data collection, annotation refinement and quality check details.

\textbf{Refinement of BEV Annotations.} As shown in \cref{Fig.UAVPTZA}, the pose of UAV is very sensitive to wind conditions, making the captured image sequence not precisely align to the reference image used for calibration. To reduce the projection error caused by this fact, we apply the image registration algorithm for all the captured images~\cite{PTZ}. The registered UAV image is illustrated in \cref{Fig.UAVPTZB}. Meanwhile, we observe that the length of vehicle in UAV's viewpoint is usually elongated due to the perspective distortion (\cref{Fig.Perspective}). Specifically, if the UAV flies at height $H$, the vehicle's real length $l$, real height $h$ and the observed length $l'$ should approximately satisfy the following triangle similarity:
\begin{equation}
\begin{split}
    \frac{l' + d}{H} &= \frac{l+d}{H-h}, \\
    l &= \frac{l'+d}{H}\left(H-h\right)-d,
\end{split}
\end{equation}
where $d$ is the projected length of the shortest line between the UAV and the vehicle \wrt the ground plane. We then use $l$ as the final box length in annotations.

\begin{figure}
    \centering
    \begin{subfigure}[b]{0.2848\linewidth}
        \centering
        \includegraphics[width=\linewidth]{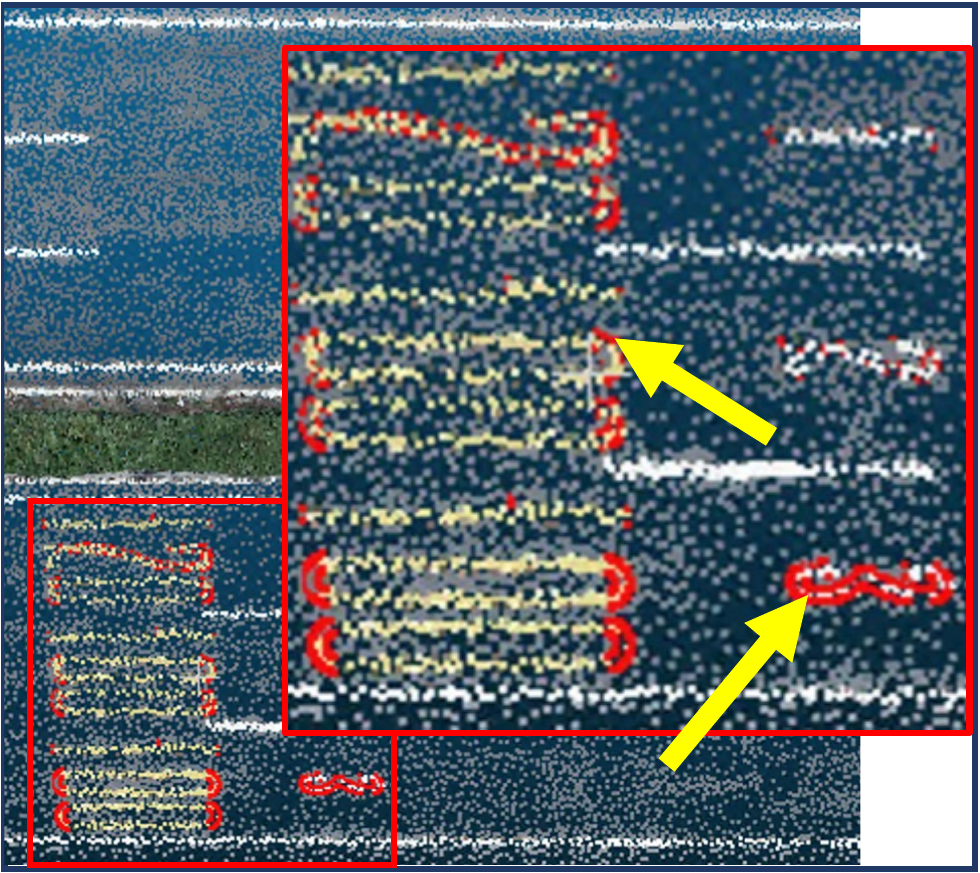}
        \caption{}
        \label{Fig.SceneError}
    \end{subfigure}
    \begin{subfigure}[b]{0.2947\linewidth}
        \centering
        \includegraphics[width=\linewidth]{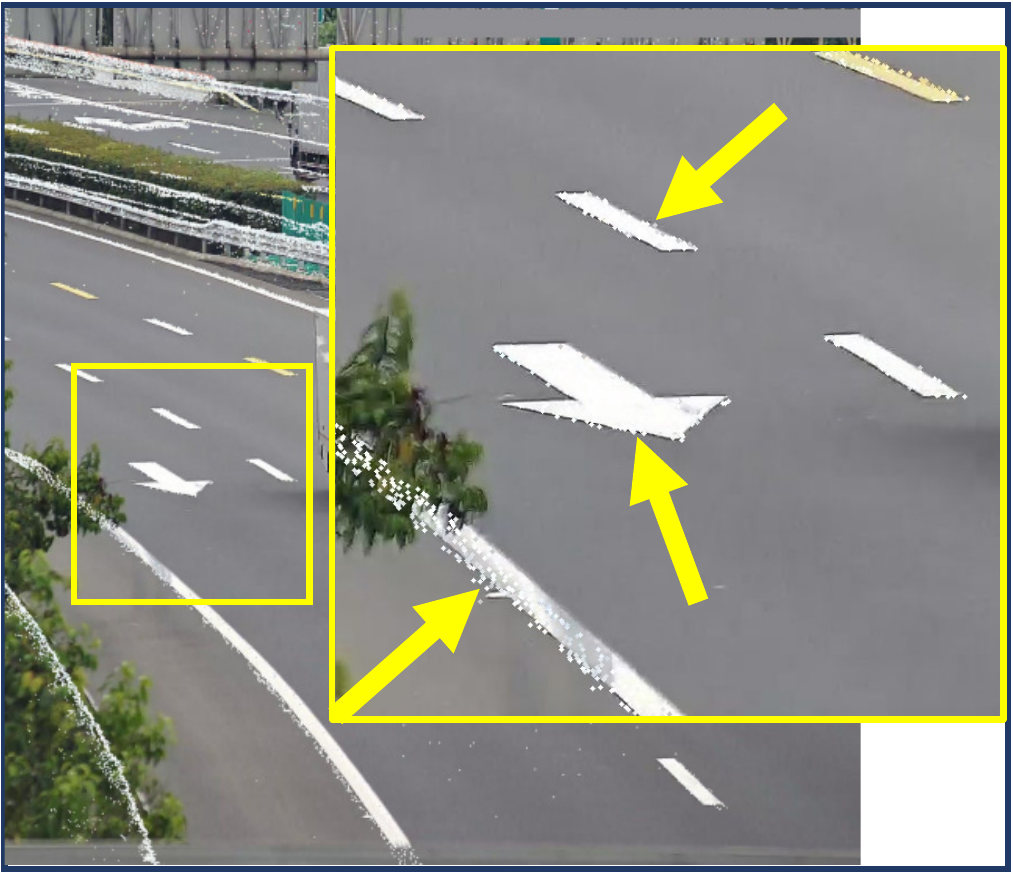}
        \caption{}
        \label{Fig.ProjectionError}
    \end{subfigure}
    \begin{subfigure}[b]{0.4004\linewidth}
        \centering
        \includegraphics[width=\linewidth]{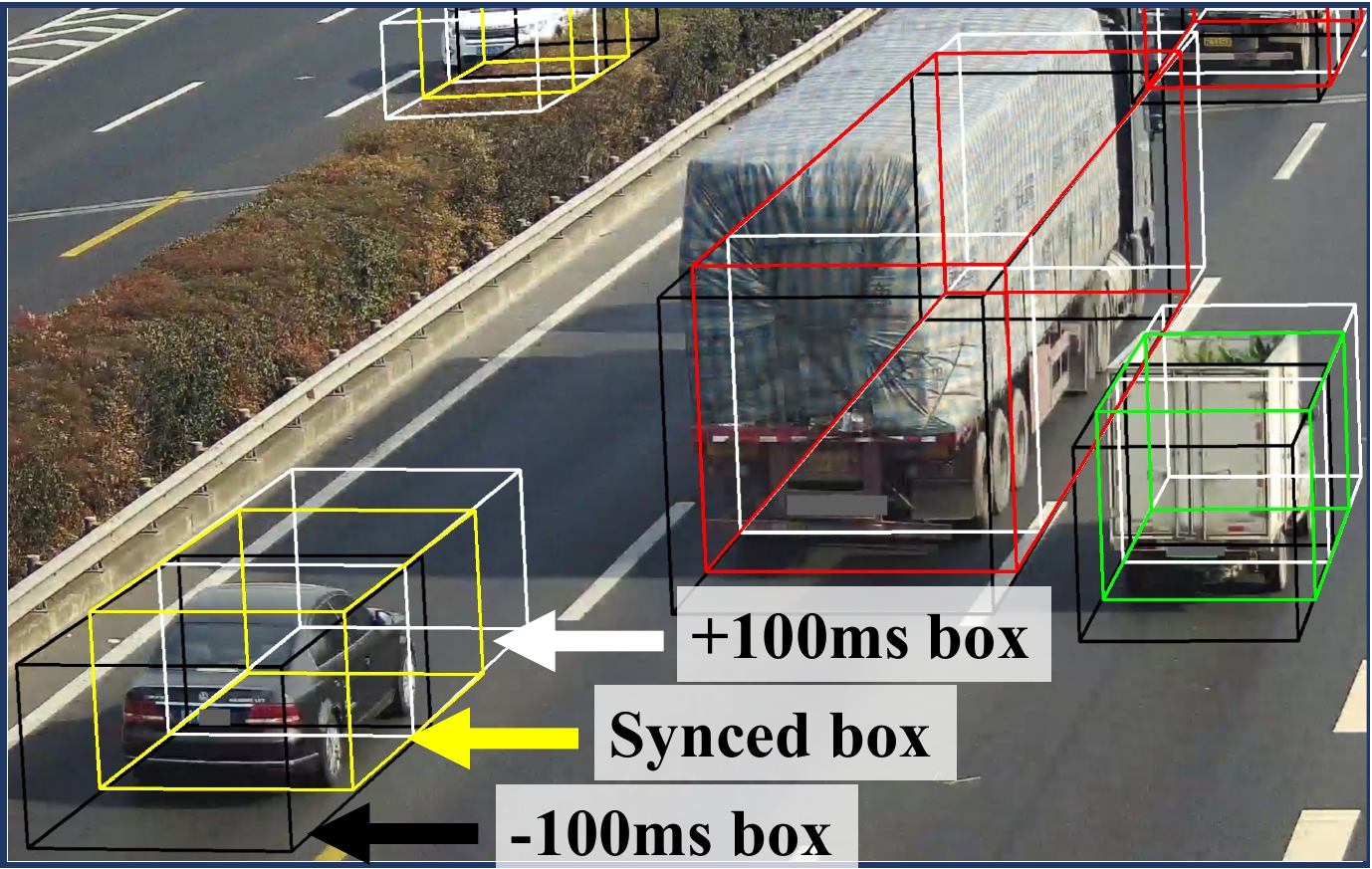}
        \caption{}
        \label{Fig.TemporalError}
    \end{subfigure}
    \caption{\textbf{(a): Static scene error visualization.} We put high-definition map as background, and plot red points sampled from 3D reconstruction as overlay. \textbf{(b): Calibration and projection error visualization.} We select a camera and pick a single frame as background, and project white points sampled from 3D reconstruction to this perspective view as overlay. \textbf{(c) Vehicles' location and height error.} To avoid temporal disalignment and height mismatch, we manually check the fitness of projected boxes with adjacent frames.}
    \vspace{-4em}
\end{figure}

\begin{figure}
    \centering
    \includegraphics[width=\linewidth]{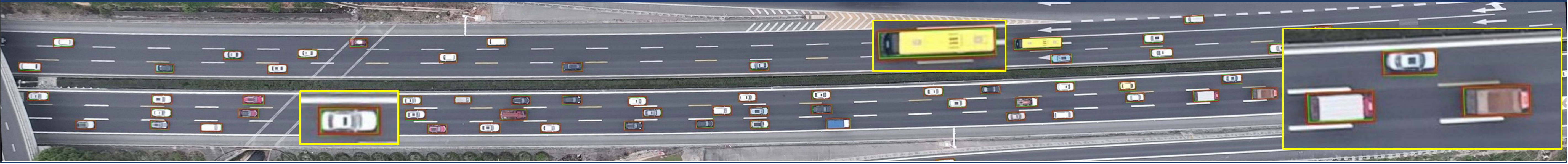}
    \caption{We visualize the vehicles' length and width error in UAV view. Green boxes indicate human annotations, while red boxes indicate model predictions.}
    \label{Fig.UAVError}
\end{figure}

\textbf{Annotation Quality Verification.} The annotation quality is influenced by following parts:

\textbf{a) Static scene error}, which is verified to be $<10cm$ as stated in main paper. Furthermore, we use the high-definition map as background and plot 3D reconstruction points as overlay to visualize the static scene error in \cref{Fig.SceneError}. As shown in the figure, all red points are precisely located to fit with the background.

\textbf{b) Projection error}. We optimize the calibration to get final projection error to be \textbf{less than 5px}. The visualization of perspective projection is shown in \cref{Fig.ProjectionError}. In the figure, background is a captured frame from a camera, while white points are sampled from 3D reconstruction and then projected to this perspective view. These points also fit well with the background.

\textbf{c) Vehicles' location and height error.} We manually check the fitness of projected boxes with adjacent frames, as shown in \cref{Fig.TemporalError}, to avoid temporal disalignment and height mismatch.

\textbf{d) Vehicles' width and length error.} With the length refinement in \cref{Fig.Perspective}, we visualize model prediction error (red boxes) with human annotations (green boxes), as shown in \cref{Fig.UAVError}. Furthermore, we pick $40k$ boxes and compare length and width with human annotations. The error is \textbf{less than 1.16px} in UAV view, corresponding to \textbf{12.53cm} in physical size.

\begin{table}[htp]
    \centering
    \caption{The mAP scores of our UAV detector. False positives can be filtered in the association stage.}
    \label{Tab.UAV-mAP}
    \begin{tabular}{@{}lcc@{}}\toprule
    Class    ~~~~~ & ~~~~Recall~~~~ & ~~~~~mAP~~~~~   \\\midrule
    Car       & $1.000$  & $0.906$ \\
    Van       & $1.000$  & $0.956$ \\
    Truck     & $1.000$  & $0.993$ \\
    Bus       & $1.000$  & $0.997$ \\\bottomrule
    \end{tabular}
\end{table}

\textbf{Details of UAV 2D Detector and Tracker.} To obtain a high-performance UAV 2D detector in our scenarios under various weather and light conditions, we collect a total of $24,484$ images at different times and locations. These images are manually annotated with a team of professional human annotators. We then fine-tune the RTMDet-L~\cite{RTMDet} model using the annotated images for $36$ epochs with the default training config (\href{https://github.com/open-mmlab/mmrotate/blob/1.x/configs/rotated_rtmdet/rotated_rtmdet_l-3x-dota_ms.py}{rotated\_rtmdet\_l-3x-dota\_ms}). The resulting model has a $96.3\%$ mAP in average in a test set of $2,450$ images under threshold IoU$= 0.5$, as reported in \cref{Tab.UAV-mAP}. We then use this model to annotate all the vehicles. Next, we perform the association for each data clip. We use the OC-SORT~\cite{OCSORT} to associate identical vehicles across frames to produce trajectories. The algorithm is configured with gIoU threshold $=0.5$, $\lambda=0.2$ and $\Delta t = 3$. After association, the generated trajectories need further refinement since a few false positives may produce wrong short trajectory and the false negatives make the whole trajectory interrupted at middle frame. We eliminate these flaws by filtering out trajectories whose duration is less than $1s$ or using linear interpolation to generate missed boxes based on history and future adjacent frames. Combined with these strategies, we reach 623 (0.0086\%) / 454 (0.0063\%) false positives / negatives and 24 (0.016\%) ID switches over 7.26M boxes, as stated in main paper. To investigate the impact of the label noise, we manually refine annotations in both training set and validation set, and further train RoBEV with / without the noise. Then, we obtain the same NDS on the refined validation set. Therefore, the label noise introduced by our annotation pipeline has negligible impact for dataset usage.

\clearpage

\clearpage

\subsection{More Statistics}
The statistic of vehicle orientation (yaw) is placed in \cref{Fig.Orientation}. The distribution of orientation concentrates around $0\degree$ and $180\degree$, corresponding to vehicles that toward the positive or negative direction along X-axis, respectively.

We also compare the annotation density with other datasets in \cref{Tab.Density}. \enquote{Car} and \enquote{Van} are categorized into small vehicles while \enquote{Bus} and \enquote{Truck} are categorized into large vehicles. We could see that the number of boxes per image in RoScenes for both types is much larger than all the compared datasets ($19.8\times$ to nuScenes, $4.9\times$ to Rope3D). The statistic shows the crowded traffic in RoScenes.

\begin{table}[t]
\centering
\caption{Performance comparison on $\mathit{o}$RoScenes. We randomly drop $1/2/3$ cameras to simulate the offline errors.}
\label{Tab.Offline}
\resizebox{\linewidth}{!}{
\begin{tabular}{@{}lrrrrrrrrrrrr@{}}
\toprule
 \multicolumn{1}{c}{\multirow{4}{*}{Method}}    & \multicolumn{4}{c}{$1$ Camera} & \multicolumn{4}{c}{$2$ Cameras} & \multicolumn{4}{c}{$3$ Cameras} \\ \cmidrule(lr){2-5}\cmidrule(lr){6-9}\cmidrule(lr){10-13}
    & \multicolumn{2}{c}{Easy} & \multicolumn{2}{c}{Hard} & \multicolumn{2}{c}{Easy} & \multicolumn{2}{c}{Hard} & \multicolumn{2}{c}{Easy} & \multicolumn{2}{c}{Hard}\\ \cmidrule(lr){2-3}\cmidrule(lr){4-5}\cmidrule(lr){6-7}\cmidrule(lr){8-9}\cmidrule(lr){10-11}\cmidrule(lr){12-13}
    & NDS & mAP & NDS & mAP & NDS & mAP & NDS & mAP & NDS & mAP & NDS & mAP \\ \midrule
 BEVDet~\cite{BEVDet}    & $0.404$ & $0.176$ & $0.371$ & $0.114$ & $0.399$ & $0.153$ & $0.369$ & $0.098$ & $0.387$ & $0.133$ & $0.361$ & $0.086$  \\
 BEVDet4D~\cite{BEVDet4D}   & $0.409$ & $0.165$ & $0.374$ & $0.110$ & $0.387$ & $0.135$ & $0.362$ & $0.092$ & $0.380$ & $0.119$ & $0.353$ & $0.080$ \\
 SOLOFusion~\cite{SOLOFusion}  &  $0.297$ & $0.109$ & $0.201$ & $0.059$ & $0.256$ & $0.063$ & $0.189$ & $0.037$ & $0.244$ & $0.056$ & $0.148$ & $0.012$ \\
 BEVFormer~\cite{BEVFormer} & $0.663$ & $0.556$ & $0.608$ & $0.438$ & $0.626$ & $0.491$ & $0.575$ & $0.381$ & $0.555$ & $0.378$ & $0.520$ & $0.299$\\\midrule
 DETR3D~\cite{DETR3D}    & $0.635$ & $0.522$ & $0.556$ & $0.366$ & $0.570$ & $0.421$ & $0.501$ & $0.288$ & $0.467$ & $0.271$ & $0.427$ & $0.197$ \\
 PETRv2~\cite{PETRv2}   & $0.581$ & $0.456$ & $0.507$ & $0.322$ & $0.524$ & $0.365$ & $0.466$ & $0.257$ & $0.357$ & $0.101$ & $0.349$ & $0.085$\\
 StreamPETR~\cite{StreamPETR}    & $0.556$ & $0.414$ & $0.457$ & $0.230$ & $0.498$ & $0.328$ & $0.415$ & $0.180$ & $0.423$ & $0.210$ & $0.358$ & $0.105$  \\
 RoBEV (Ours) & $0.654$ & $0.539$ & $0.618$ & $0.449$ & $0.608$ & $0.463$ & $0.581$ & $0.386$ & $0.542$ & $0.357$ & $0.523$ & $0.302$\\
\bottomrule
\end{tabular}
}
\end{table}

\subsection{Robustness Evaluation}
In the real-world roadside systems, the cameras usually suffer from the shaking problems due to the wind effect and big-vehicle passing. Therefore, we take an evaluation to study the robustness of all the BEV detection models. The models are tested without extra tuning under the following two types of settings:

\noindent\textbf{$\bm{\mathit{o}}$RoScenes} randomly drops $1\!\sim\!3$ views for every clip in test set to simulate exceptional camera offline. The result is shown in \cref{Tab.Offline}. When we only discard a single view, all methods have averagely $\sim\!10\%$ lower performance than the original. When the number of offline cameras increases, their performance drop significantly, especially the mAP metric. It is reasonable since the absence of a few cameras will lead to the occurrence of blind zones, making the models hard to locate highly-occluded objects with the remaining available views.

\noindent\textbf{$\bm{\mathit{p}}$RoScenes} randomly imposes the perspective perturbation for all images. The perturbation performs random pan$\sim\mathcal{N}\left(0, 3.33\right)$, tilt$\sim\mathcal{N}\left(0, 1.67\right)$ and zoom $\sim\mathcal{N}\left(1.0, 0.03\right)$ which simulates the shaking of cameras. The performance under perturbation is placed in the left part of \cref{Tab.Perturb} (\textit{w/o} Registration). Such a small perturbation makes all methods have $\sim\!20\%$ performance drop. We then use the image registration method (the same as \cref{Sec.Refine}) to rectify the inputs and compare the performance in the right part (\textit{w/} Registration), which largely alleviates the performance downgrade (only $\sim\!2\%$ performance drop).

\begin{table}[t]
\centering
\caption{Performance comparison on $\mathit{p}$RoScenes. The random perspective perturbation are applied on images.}
\label{Tab.Perturb}
\begin{tabular}{@{}lrrrrrrrr@{}}
\toprule
 \multicolumn{1}{c}{\multirow{4}{*}{Method}}    & \multicolumn{4}{c}{\textit{w/o} Registration} & \multicolumn{4}{c}{\textit{w/} Registration} \\ \cmidrule(lr){2-5}\cmidrule(lr){6-9}
    &\multicolumn{2}{c}{Easy}&\multicolumn{2}{c}{Hard}&\multicolumn{2}{c}{Easy}&\multicolumn{2}{c}{Hard}\\\cmidrule(lr){2-3}\cmidrule(lr){4-5}\cmidrule(lr){6-7}\cmidrule(lr){8-9}
    & NDS & mAP & NDS & mAP & NDS & mAP & NDS & mAP\\ \midrule
 BEVDet~\cite{BEVDet}  & $0.348$ & $0.080$ & $0.337$ & $0.058$ & $0.497$ & $0.292$ & $0.428$ & $0.181$    \\
 BEVDet4D~\cite{BEVDet4D} & $0.348$ & $0.080$ & $0.340$ & $0.064$ & $0.418$ & $0.192$ & $0.377$ & $0.113$ \\
 SOLOFusion~\cite{SOLOFusion} & $0.280$ & $0.108$ & $0.195$ & $0.064$ & $0.295$ & $0.117$ & $0.201$ & $0.066$  \\
 BEVFormer~\cite{BEVFormer} & $0.445$ & $0.233$ & $0.433$ & $0.197$ & $0.685$ & $0.598$ & $0.584$ & $0.419$  \\\midrule
 DETR3D~\cite{DETR3D}  & $0.396$ & $0.153$ & $0.379$ & $0.116$  & $0.717$ & $0.641$ & $0.627$ & $0.467$  \\
 PETRv2~\cite{PETRv2}  & $0.391$ & $0.137$ & $0.381$ & $0.114$ & $0.661$ & $0.578$ & $0.575$ & $0.405$     \\
 StreamPETR~\cite{StreamPETR}  & $0.428$ & $0.212$ & $0.381$ & $0.117$ & $0.607$ & $0.492$ & $0.481$ & $0.265$   \\
RoBEV (Ours)  & $0.441$ & $0.225$ & $0.435$ & $0.199$ & $0.745$ & $0.677$ & $0.666$ & $0.512$  \\
\bottomrule
\end{tabular}
\end{table}

\begin{figure}[h]
    \centering
     \begin{subfigure}[b]{0.49\linewidth}
         \centering
         \includegraphics[width=\linewidth]{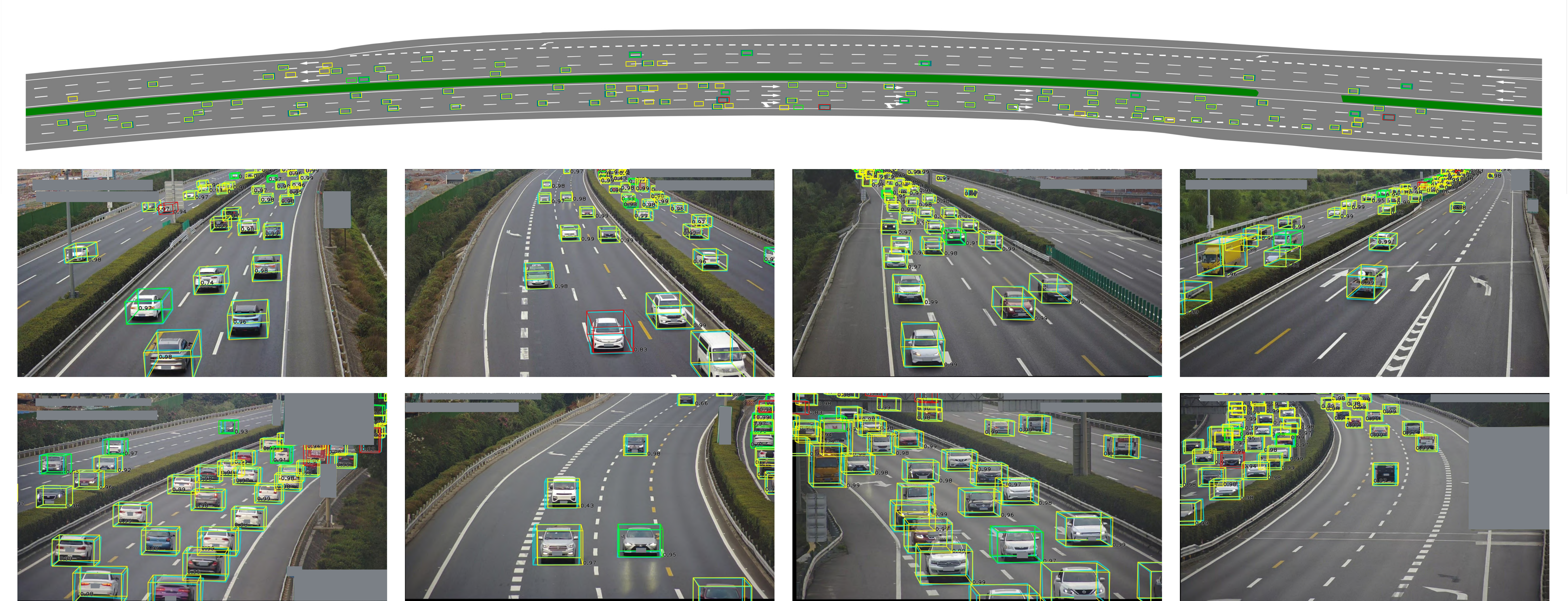}
         \caption{Detection on RoScenes test set.}
         \label{Fig.original}
     \end{subfigure}
     \begin{subfigure}[b]{0.49\linewidth}
         \centering
         \includegraphics[width=\linewidth]{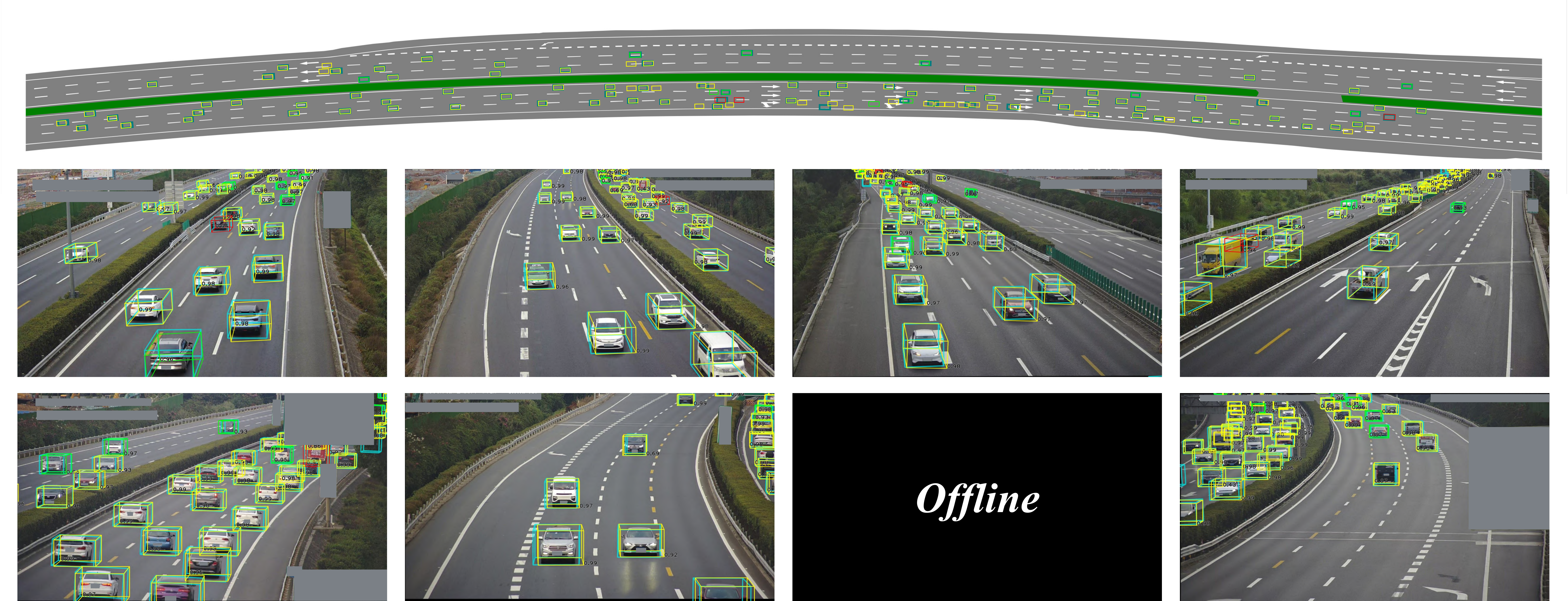}
         \caption{Detection on $\mathit{o}$RoScenes with \textbf{1} offline cam.}
         \label{Fig.Off1}
     \end{subfigure}
     \begin{subfigure}[b]{0.49\linewidth}
         \centering
         \includegraphics[width=\linewidth]{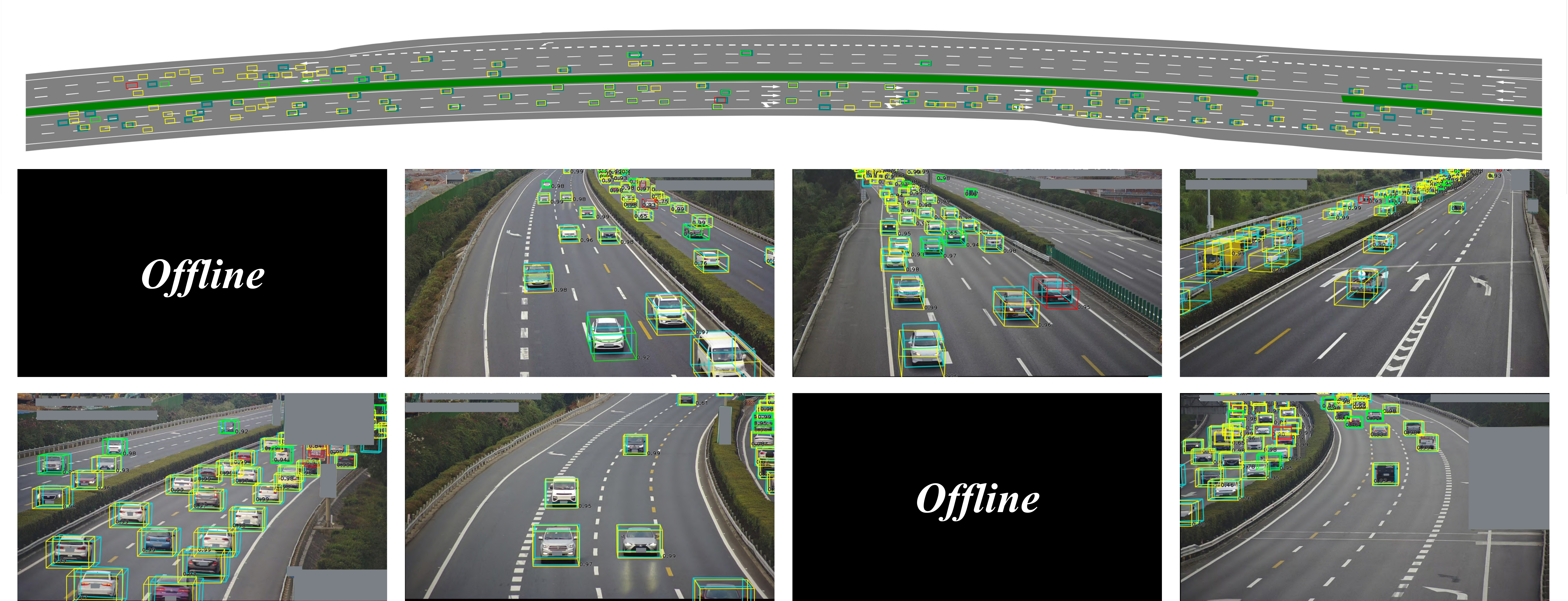}
         \caption{Detection on $\mathit{o}$RoScenes with \textbf{2} offline cams.}
         \label{Fig.Off2}
     \end{subfigure}
     \begin{subfigure}[b]{0.49\linewidth}
         \centering
         \includegraphics[width=\linewidth]{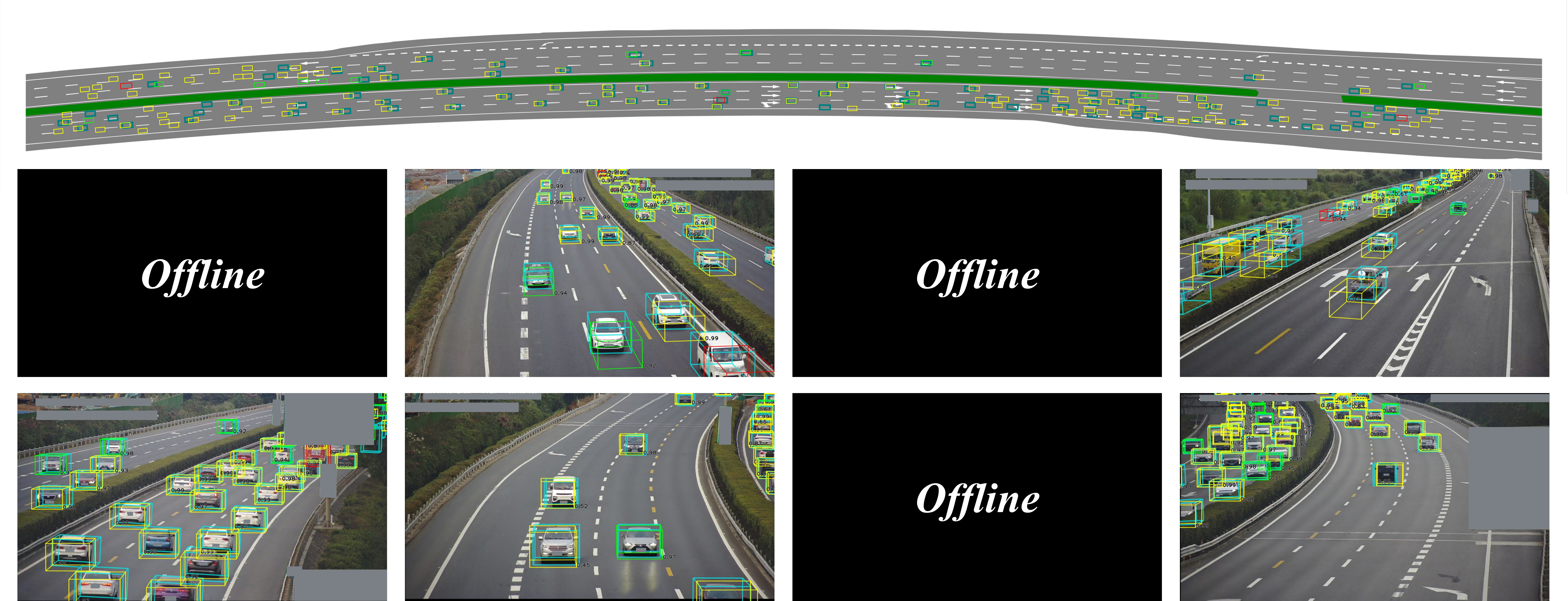}
         \caption{Detection on $\mathit{o}$RoScenes with \textbf{3} offline cams.}
         \label{Fig.Off3}
     \end{subfigure}
     \begin{subfigure}[b]{0.49\linewidth}
         \centering
         \includegraphics[width=\linewidth]{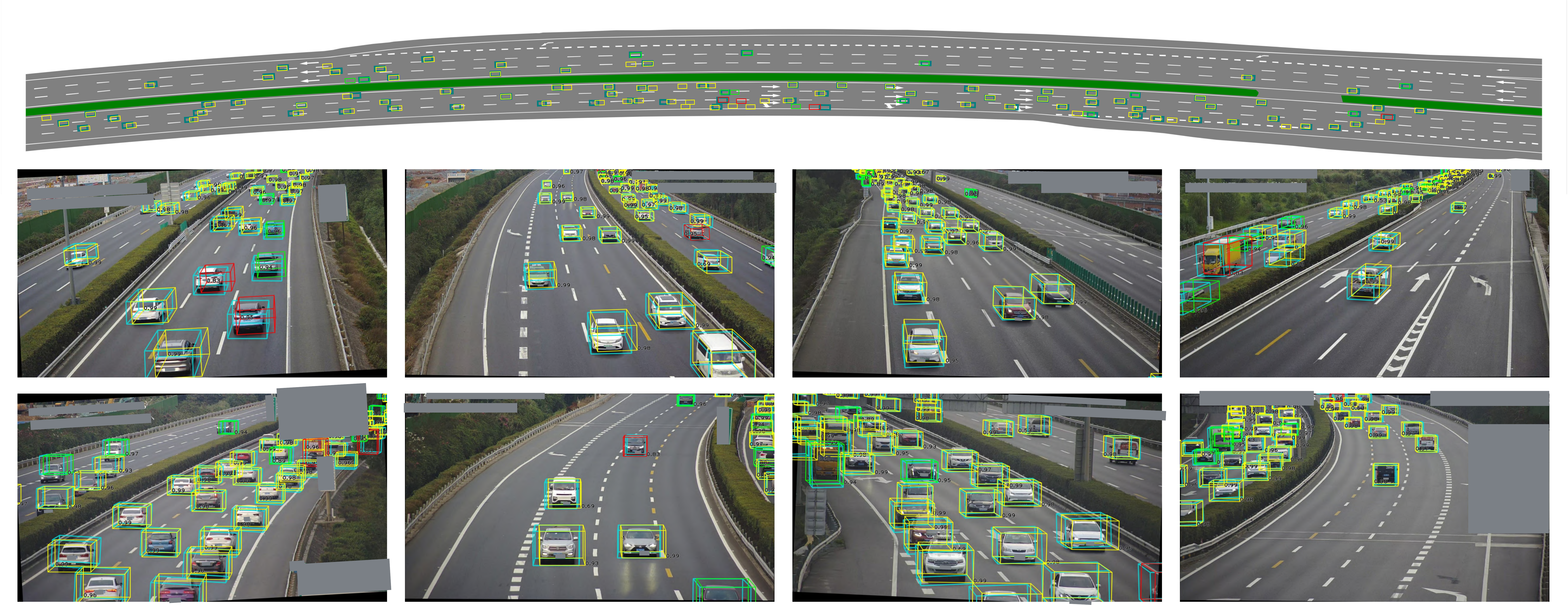}
         \caption{Detection on $\mathit{p}$RoScenes.}
         \label{Fig.Perturb}
     \end{subfigure}
     \begin{subfigure}[b]{0.49\linewidth}
         \centering
         \includegraphics[width=\linewidth]{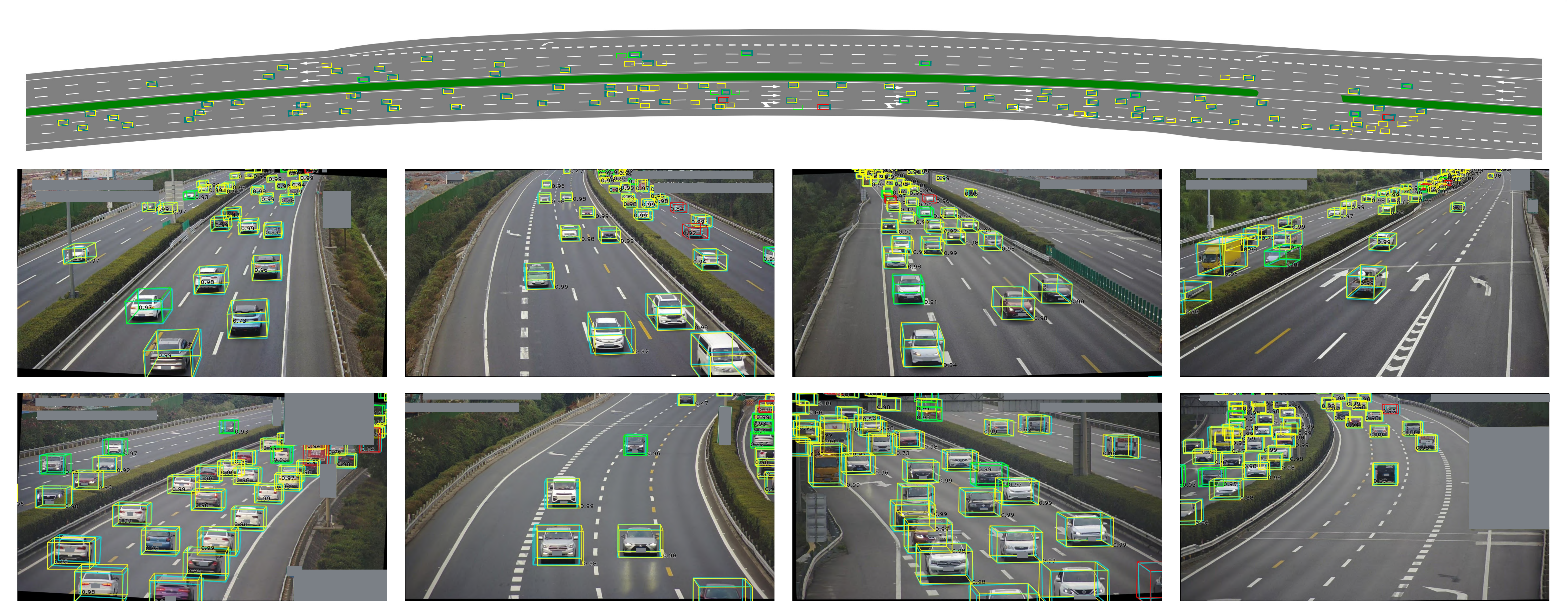}
         \caption{Detection on $\mathit{p}$RoScenes with rectified input.}
         \label{Fig.PTZ}
     \end{subfigure}
     \caption{Visualization of detection results for RoBEV. Predictions are shown in yellow, green, blue, red for car, van, bus, and truck. Groundtruths are shown in cyan.}
     \label{Fig.Vis}
\end{figure}

From the above tests, current methods are still not feasible when facing noisy or novel inputs without adaptions or adjustments. A future study on these topics would be valuable for real-world roadside perception systems.

\section{Additional Experiments Analysis}

\subsection{Additional Notes}
\label{Sec.Additional}

\textbf{Implementing RoBEV.} The detailed steps to obtain feature-guided position embedding are listed as follows: 1) A two-layer convolutional network projects image features $\bm{\mathcal{I}}$ to a 1D feature map $\bm{\mathcal{D}} \subseteq \bm{\mathbb{R}}^{\left|C\right|\times H \times W \times 1}$. 2) We concatenate $\bm{\mathcal{D}}$ with $\bm{\mathcal{M}}$ to get 3D points, which are then projected to BEV 3D space as $\bm{\mathcal{P}} = t\left( \bm{\mathcal{I}}, \bm{\mathcal{M}}\right) \subseteq \bm{\mathbb{R}}^{\left|C\right|\times H \times W \times 3}$ by camera intrinsic and extrinsic parameters. 3) A learnable 3D positional encoding~\cite{LPE} $\varphi: \mathbb{R}^3\rightarrow\mathbb{R}^D$ converts $\bm{\mathcal{P}}$ to $D$-dim positional embedding. Note that $\bm{q}$ is obtained by converting learnable 3D BEV reference points to $D$-dim $\bm{q} = \varphi\left(\mathit{ref}\right), \mathit{ref}\subseteq \mathbb{R}^{3}$, which is in the same way as 3) does, therefore we set $\varphi$ to be shared for two conversions.

\noindent\textbf{Discussion.} PETR series included in this benchmark have implemented the squeeze-and-excitation operation over static position embedding based on image input. Therefore, their models actually introduce implicit vision prior in position embedding~\cite{PETRv2}. However, we do not observe a noticeable performance gain from this operation.

\noindent\textbf{Performance of Other Methods.} We have tried to evaluate two other methods for BEV perception: BEVHeight~\cite{BEVHeight}\footnote{We modify BEVHeight to accept multi-view images as input for supporting BEV perception.} and SparseBEV~\cite{SparseBEV} on RoScenes. However, both of them do not achieve reasonable performance (NDS: BEVHeight = 0.279, SparseBEV = 0.256). Specifically, the explicit method BEVHeight estimates vehicle height and use similar triangles to derive vehicles's world coordinate. However, BEVHeight uses the ground plane hypothesis to perform derivation but the RoScenes' ground over the whole perception cuboid is not a strict plane, thus introduces errors in height estimation. Implicit SparseBEV follows the DETR3D framework and incorporates with multi-scale features and temporal fusion, but the fusion requires velocity estimation for object alignment, which may suffer the convergence issue in training in our dataset. We will leave them for future study to demystify these issues.


\begin{figure}[t]
    \centering
    \includegraphics[width=0.4\linewidth]{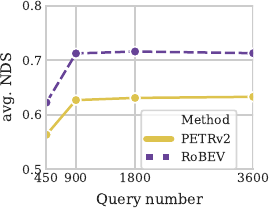}
    \caption{\textit{avg.} NDS \wrt number of queries comparison between PETRv2 and ours.}
    \label{Fig.QueryNum}
\end{figure}

\begin{table}[t]
    \centering
    \caption{Performance comparison with PETRv2 on nuScenes.}
    \label{Tab.nuScenes}
    \begin{tabular}{@{}lrrrrrrr@{}} \toprule
    Method  & NDS & mAP & mATE & mASE & mAOE & mAVE & mAAE \\\midrule
    PETRv2  & $0.503$ & $0.410$ & $0.723$ & $0.269$ & $0.453$ & $0.389$ & $0.193$ \\
    RoBEV & $0.505$ & $0.412$ & $0.718$ & $0.266$ & $0.451$ & $0.387$ & $0.192$ \\\bottomrule
    \end{tabular}
\end{table}

\subsection{Additional Experiments on RoBEV}

\textbf{Impact of Query Quantity.} We vary the number of detection queries from $450$ to $3,600$ for PETRv2 and RoBEV and report results in the left part of \cref{Fig.QueryNum}. Both methods show relatively low performance when using only $450$ queries and receive $7.7\%$ performance improvement in average when the query number increases to $900$. However, there is no noticeable gain when we further increase the query number.

\noindent\textbf{Visualization.} The detection results of our RoBEV are visualized in \cref{Fig.Vis} with samples from normal RoScenes test set as well as the $\mathit{o}$RoScenes, $\mathit{p}$RoScenes test sets. In normal RoScenes test set (\cref{Fig.original}), RoBEV could precisely detect most of appeared vehicles with correct location, size, orientation. When tested in $\mathit{o}$RoScenes (\cref{Fig.Off1,Fig.Off2,Fig.Off3}), a lot of false positive detections appear in the region covered by offline cameras. In $\mathit{p}$RoScenes (\cref{Fig.Perturb}), the predicted boxes are wrongly located and make translation error high. But this is immediately fixed by applying image registration on the perturbed images (\cref{Fig.PTZ}).

\noindent\textbf{Result on nuScenes.} Our RoBEV is applicable in vehicle-side BEV perception task. Therefore, we test its performance in the nuScenes dataset. We adopt Res-50 with deformable convolution as backbone, and train on the nuScenes training set with input size $[320, 800]$ for $24$ epochs. The comparison with PETRv2 is shown in \cref{Tab.nuScenes}. Actually, our method does not receive noticeable performance gain under this dataset. The main reason is that nuScenes only contains a single camera layout, therefore a static position embedding used in PETRv2 is enough to encode spatial information into 2D features.

\subsection{Monocular 3D Object Detection Task}

Since the single camera setting is often used in the industrial applications due to its affordability, thereby, it is crucial to examine the effectiveness of monocular 3D object detection in our RoScenes dataset. To enhance the efficiency of the experimental process, we adopt a random selection approach to partition our RoScenes dataset, reserving one-fourth of the entire dataset.
This subset encompasses 32,000 images for training purposes, while an additional set of 8,000 images are allocated for validation. Objects with occlusion $\mathit{occ} < 0.8$ are easy samples, while the remaining are hard.
By employing this sampling strategy, we ensure a representative and substantial dataset for comprehensive evaluation and analysis.

\noindent\textbf{Implementation Details.} We compare performance of the following three models:

a) Monoflex~\cite{zhang2021objects} decouples the features learning of truncated objects with backbone DLA-34~\cite{yu2018deep}.

b) MonoDETR~\cite{zhang2023monodetr} predicts the pixel-level depth estimation and concatenates it with the image features to generate the final results with DETR~\cite{DETR} backbone.

c) BEVHeight~\cite{BEVHeight} estimates the distance between vehicle and camera via calculating the ratio of vehicle's height to reference height.

c) PDR~\cite{sheng2023pdr} proposes an improved perspective projection-based depth generation method.

All models are trained using the AdamW optimizer~\cite{adamW} on two Tesla V100 GPUs, employing a batch size of 24 with 100 epochs with their official released codes. The original image is padded and downsampled to a resolution of $384\times 640$. All methods solely employ horizontal random flipping as their data augmentation strategy. The maximum depth value is set to 800m. During the testing phase, bounding boxes with a confidence score threshold of 0.1 or higher are considered. The maximum number of detectable objects per image is limited to 300. Each model provides the outputs of object's location, size, heading angle, and pitch angle. Consequently, the coordinates of the corresponding eight corner points can be calculated. The pytorch3d toolbox~\cite{pytorch3d} is utilized to compute the 3D Intersection over Union (IoU) between the predicted and ground-truth objects, utilizing their respective eight corner points.

Inspired by KITTI~\cite{KITTI} and Rope3D~\cite{Rope3D} datasets, we adopt the 40-point Interpolated Average Precision metric~\cite{simonelli2019disentangling} ($\text{AP}_{40}$) to perform fair comparison. All the results are reported under IoU = 0.3 and 0.5, respectively. All categories are trained within a single model. 


\begin{table}[h!]
\centering
\caption{Performance comparison of different monocular 3D object detectors on our RoScenes dataset with IoU = 0.3 and 0.5.}
\label{table:mono_all_data}
\begin{tabular}{lrrrr}\toprule
\multirow{2}{*}{Method} & \multicolumn{2}{c}{Easy} & \multicolumn{2}{c}{Hard}  \\ \cmidrule(lr){2-3}\cmidrule(lr){4-5}
&  IoU=0.3 & IoU=0.5 & IoU=0.3 & IoU=0.5  \\\midrule
MonoFlex~\cite{zhang2021objects}  & $19.29$ & $9.36$ & $1.71$ & $0.72$ \\
MonoDETR~\cite{zhang2023monodetr} & $0.21$ & $0.04$ & $0.02$ & $0.01$ \\
BEVHeight~\cite{BEVHeight} & $21.20$ & $12.55$  & $2.12$  & $0.98$  \\
PDR~\cite{sheng2023pdr} & $25.90$ & $17.15$  & $2.89$ & $1.53$ \\\bottomrule
\end{tabular}
\end{table}
\begin{figure}[h!]
\centering
    \includegraphics[width=\linewidth]{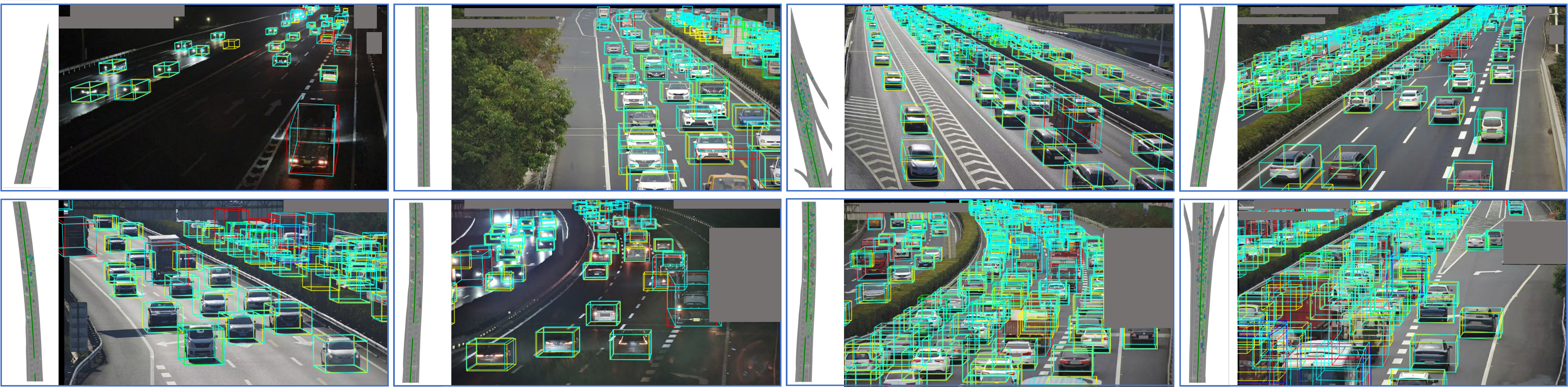}
    \caption{Monocular 3D object detection results. The visualization results are presented for both the camera view and the Bird's Eye View. Predictions are shown in yellow, green, blue, red for car, van, bus, and truck.
    Groundtruths are shown in cyan.}
    \label{Fig.PDR_vis}
\end{figure}
\begin{table}[h!]
    \centering
    \caption{Performance comparison on different categories with IoU=0.3 and 0.5, respectively.}
    \label{table:mono_classes}
    \resizebox{\linewidth}{!}{
    \begin{tabular}{lcccccccc}\toprule
\multirow{2}{*}{Method} & \multicolumn{2}{c}{car} & \multicolumn{2}{c}{van} & \multicolumn{2}{c}{bus} & \multicolumn{2}{c}{truck}  \\ \cmidrule(lr){2-3}\cmidrule(lr){4-5}\cmidrule(lr){6-7}\cmidrule(lr){8-9}
& Easy & Hard & Easy & Hard & Easy & Hard & Easy & Hard  \\\midrule
MonoFlex~\cite{zhang2021objects}  & 19.29 / 9.36 & 1.71 / 0.72 & 1.23 / 0.62 & 0.18 / 0.09 & 0.09 / 0.05 & 0.01 / 0.00 & 2.13 / 0.92 & 0.26 / 0.11\\
BEVHeight~\cite{BEVHeight} & 21.20 / 12.55 &  2.12 / 0.98   &  1.54 / 0.77  & 0.18 / 0.11     &    0.12 / 0.09   &  0.01 / 0.00    &   2.15 / 1.01   & 0.27 / 0.12        \\
PDR~\cite{sheng2023pdr} & 25.90 / 17.15  & 2.89 / 1.53 & 1.70 / 0.82 & 0.20 / 0.14 & 0.21 / 0.13 & 0.02 / 0.01 & 2.16 / 1.13 & 0.26 / 0.12\\\bottomrule
\end{tabular}
}
\end{table}

\subsection{Main Result}
The performances of monocular 3D detection on the RoSecenes dataset are depicted in Table~\ref{table:mono_all_data}. Under the easy level and IoU=0.3, the best performance is achieved by PDR~\cite{sheng2023pdr} with $25.90\% \text{AP}_{40}$. In general, our RoScenes dataset offers ample opportunities for further exploration in the field of monocular 3D object detection. In addition, MonoDETR has relatively low performance. The pixel-wise depth prediction employed in this method exhibits limitations in robustly handling various camera layouts.  In this way, MonoDETR has difficulty handling large ranges of depth values.  In contrast, MonoFlex and PDR, which are perspective projection-based methods, indirectly predict object depth values by estimating both the 3D physical height and 2D projected height. Similarly, BEVHeight also performs indirect depth estimation and achieves reasonable performance. On the contrary, MonoDETR achieves near-zero values for all metrics. In a nutshell, these indirect depth prediction methods exhibit significantly superior performance on RoScenes.

\noindent\textbf{Performance Across Different Categories.} For reference, we present the detailed 3D object detection performance for different categories in Table~\ref{table:mono_classes}. Since cars constitute 80\% of the overall objects, the detection performance for the car category exhibits the best.


\noindent\textbf{Visualization.} In Figure~\ref{Fig.PDR_vis}, we provide the visualization results of monocular 3D object detection using PDR method.

\end{document}